\documentclass[5p]{elsarticle}

\usepackage{graphicx}
\usepackage{caption}
\usepackage{subcaption}
\usepackage{amsmath}
\usepackage{algorithmic}
\usepackage{algorithm}
\usepackage{lineno}
\usepackage{multirow}
\usepackage{tabularx}
\usepackage{booktabs}
\usepackage{url}
\usepackage{color}
\usepackage[figuresright]{rotating}
\usepackage[utf8]{inputenc}
\usepackage[T1]{fontenc}

\newcolumntype{Y}{>{\arraybackslash}X}

\journal{Pattern Recognition}

\begin{document}

\begin{frontmatter}

\title{Potential Anchoring for imbalanced data classification}

\author[agh]{Micha\l{} Koziarski\corref{cor1}}
\ead{michal.koziarski@agh.edu.pl}
\cortext[cor1]{Corresponding author}

\address[agh]{Department of Electronics, AGH University of Science and Technology, Al. Mickiewicza 30, 30-059 Krak\'ow, Poland}

\begin{abstract}
Data imbalance remains one of the factors negatively affecting the performance of contemporary machine learning algorithms. One of the most common approaches to reducing the negative impact of data imbalance is preprocessing the original dataset with data-level strategies. In this paper we propose a unified framework for imbalanced data over- and undersampling. The proposed approach utilizes radial basis functions to preserve the original shape of the underlying class distributions during the resampling process. This is done by optimizing the positions of generated synthetic observations with respect to the potential resemblance loss. The final Potential Anchoring algorithm combines over- and undersampling within the proposed framework. The results of the experiments conducted on 60 imbalanced datasets show outperformance of Potential Anchoring over state-of-the-art resampling algorithms, including previously proposed methods that utilize radial basis functions to model class potential. Furthermore, the results of the analysis based on the proposed data complexity index show that Potential Anchoring is particularly well suited for handling naturally complex (i.e. not affected by the presence of noise) datasets.
\end{abstract}


\end{frontmatter}

%\linenumbers

\section{Introduction}

Data imbalance \cite{Sun:2009,Krawczyk:2016,Branco:2016} occurs in the classification problem domain whenever the number of observations from one of the classes (majority class) is higher than the number of observations from one of the other classes (minority class). Existing learning algorithms are typically susceptible to the presence of data imbalance, displaying bias towards the majority class. The negative impact of data imbalance on the classifiers performance is further exacerbated by inherent data difficulty factors such as class overlap, small disjuncts, presence of noise, and insufficient number of training observations \cite{jo2004class,chen2008fast,he2009learning,stefanowski2016dealing}. 

The problem of data imbalance is ubiquitous in practical applications, affecting domains such as cancer malignancy grading \cite{koziarski2018convolutional}, industrial systems monitoring \cite{ramentol2016fuzzy}, fraud detection \cite{wei2013effective}, behavioral analysis \cite{azaria2014behavioral} and cheminformatics \cite{czarnecki2015compounds}. Furthermore, data imbalance typically leads to the more costly type of error, for instance by inducing false negatives in the medical problem domain. Because of that, imbalanced data classification remains focus of intense scientific effort. 

One of the most prevalent approaches for dealing with data imbalance are the data-level algorithms: methods that reduce imbalance either by creating new minority class observations (oversampling) or reducing the number of majority class observations (undersampling). In particular in the case of oversampling, this usually requires generation of synthetic observations to prevent overfitting. Existing oversampling strategies typically modify class distribution, focusing the process of generation of observations in specific regions based on the adapted strategy \cite{han2005borderline,Bunkhumpornpat:2009,he2008adasyn,barua2012mwmote,koziarski2019radial}. 

In this paper we propose a novel approach to imbalanced data resampling that is based on the idea of preserving the shape of the original class distributions. The proposed approach frames the resampling problem as an optimization of positions of generated observations with respect to the potential resemblance loss, a tool for evaluating the relative shape of class distributions. The main contributions of this paper can be summarized as follows:

\begin{itemize}
    \item Proposition of potential resemblance loss, which utilizes radial basis functions to evaluate the relative distribution shape for two sets of observations.
    \item Integration of potential resemblance loss into a unified over- and undersampling framework.
    \item Proposition of data difficulty index, a function measuring the complexity of the considered dataset.
    \item Experimental comparison of the proposed approach with state-of-the-art resampling strategies.
    \item Examination of factors influencing the relative performance of the proposed approach.
\end{itemize}

The rest of this paper is organized as follows. In Section~\ref{sec:related-work} we discuss the relevant scientific contributions. In Section~\ref{sec:potential-anchoring} we introduce the concept of potential resemblance loss, and discuss how it can be utilized during imbalanced data resampling. In Section~\ref{sec:data-difficulty-index} we introduce the concept of data difficulty index, a measure of dataset complexity that will be later utilized to identify the areas of applicability of the proposed algorithm. In Section~\ref{sec:experimental-study} we describe the conducted experimental study and the observed results. Finally, in Section~\ref{sec:conclusions} we present our conclusions.

\section{Related work}
\label{sec:related-work}

Two main approaches to dealing with data imbalance can be distinguished. First of all, data-level methods, the aim of which is to modify the training data to artificially reduce the degree of imbalance, either by creation of new minority class observations (oversampling) or by removal of the majority class observations (undersampling). 

By far the most prevalent oversampling paradigm are neighborhood-based algorithms originating from the Synthetic Minority Over-sampling Technique (SMOTE) \cite{chawla2002smote}. SMOTE is considered a cornerstone for the contemporary imbalanced data resampling \cite{fernandez2018smote}, having inspired numerous other approaches based on the idea of interpolating nearest minority class observations during the oversampling. These methods typically focus the oversampling in a specific regions, modifying the original class distribution in the process. Notable examples include Borderline-SMOTE (Bord) \cite{han2005borderline}, Safe-Level-SMOTE \cite{Bunkhumpornpat:2009},  Adaptive Synthetic Sampling (ADASYN) \cite{he2008adasyn}, and MWMOTE \cite{barua2012mwmote}.

However, despite their prevalence, SMOTE-based techniques have several shortcomings that limit their usefulness on datasets affected by various data difficulty factors. The most significant drawbacks of SMOTE are: that it assumes a homogeneous minority class clusters, and that it does not consider the majority objects during the resampling process. Numerous attempts have been made to address the aforementioned issues, leading to development of novel oversampling approaches \cite{Perez-Ortiz:2016,Bellinger:2018}. It was also demonstrated that SMOTE performs poorly on highly dimensional data \cite{lusa2012evaluation,maldonado2019alternative}.

One of the categories of undersampling algorithms are cleaning strategies, the aim of which is removal of a subset of existing majority class observations. This is typically done based on a heuristic strategy of finding the observations inconsistent with the remainder of the data. Early examples of cleaning strategies include methods such as Tomek Links \cite{tomek1976two} and Edited Nearest-Neighbor rule \cite{wilson1972asymptotic}. These methods are often used in combination with SMOTE oversampling. Some more recent cleaning strategies can also be distinguished. For instance, Anand et al. \cite{anand2010approach} proposed sorting the undersampled observations based on the weighted Euclidean distance from the positive samples. Smith et al. \cite{smith2014instance}  advocated for using the instance hardness criterion to determine the order of observation removal.

Clustering-based algorithms constitute another popular family of undersampling algorithms. This type of methods employs clustering, either to replace the original data with a completely novel set of observations \cite{lin2017clustering}, or to determine the most representative subset of original observations \cite{yen2009cluster}. Clustering-based undersampling was also successfully used to form classifier ensembles \cite{liu2008exploratory}, an idea that was further extended in form of evolutionary undersampling \cite{galar2013eusboost} and boosting \cite{lu2017adaptive}.

It is also worth mentioning that some data-level approaches utilizing radial basis functions can be distinguished in the literature. SWIM framework \cite{bellinger2019framework} used them to model the density of well-sampled majority class observations in the case of extreme imbalance, with the goal of constraining the oversampling regions. Radial-Based Oversampling (RBO) \cite{koziarski2019radial}, on the other hand, used radial basis functions to guide oversampling towards regions of low absolute potential, which can be interpreted as placed close to the decision border. This concept was further extended to the undersampling setting with Radial-Based Undersampling (RBU) \cite{koziarski2020radial}, where it determined the order of observation removal.

The differences between over- and undersampling algorithms have been examined in the literature, both from the theoretical, as well as the experimental standpoint. It was recognized early that over- and undersampling, despite both being a data-level approaches, face unique challenges. In particular, while the undersampling strategies are mostly concerned with elimination of information loss due to the removal of observations, the main concern for oversampling algorithms is reducing the impact of overfitting \cite{barandela2004imbalanced}.

Furthermore, some studies concerned with the question of which of the algorithms yields better performance can be distinguished. Notably, some experimental results seem to indicate that undersampling can be a preferable approach for data affected by high levels of noise \cite{van2009knowledge}. Another study concludes that an important factor affecting the relative performance is the level of imbalance, and that oversampling tends to perform better on a severely imbalanced datasets \cite{garcia2012effectiveness}. Finally, in a recent study it was demonstrated that combining over- and undersampling with a properly chosen ratio can be beneficial to the final performance \cite{koziarski2020csmoute}. However, there does not seem to be a clear consensus on which family of the algorithms outperforms the other, and in practice an experimental evaluation is usually required to determine the approach optimal for a given dataset.

Finally, the second category of methods for handling data imbalance constitutes of algorithm-level strategies. This type of methods modifies the training procedure of traditional classification algorithms to better account for the data imbalance, and to reduce its negative impact on the minority class performance. Notable examples of algorithm-level solutions include: kernel functions \cite{Mathew:2018}, splitting criteria in decision trees \cite{Li:2018}, and modifications of the underlying loss function to make it cost-sensitive \cite{Khan:2018}. However, contrary to the data-level approaches, algorithm-level strategies require a specific choice of classification algorithm, making them less flexible than data-level approaches. Still, in many cases, they are reported to lead to a better performance \cite{Fernandez:2018}.

\section{Potential Anchoring}
\label{sec:potential-anchoring}

In this paper we propose a novel, unified framework for imbalanced data over- and undersampling. The proposed framework utilizes radial basis functions to measure the resemblance of class distributions between the original and synthesized observations. In the remainder of this section we outline the motivation behind the proposed approach, introduce the concept of potential resemblance loss, and describe how can it be integrated into both over- and undersampling procedure.

\subsection{Motivation}

Early attempts at dealing with data imbalance utilized random oversampling, a process during which identical copies of the existing observations are created. However, it was soon realized that creation of exact duplicates of the existing observations can lead to overfitting of certain classifiers, which motivated the proposal of SMOTE \cite{chawla2002smote}. Contrary to the exact duplication of the existing observations, SMOTE creates synthetic observations by interpolating the existing ones. While this process was empirically shown to outperform random oversampling, it can alter the produced minority class distribution. This is due to the fact the density of the original observations is not guaranteed to be identical to that of the generated ones.

Since its inception SMOTE became a cornerstone for the contemporary imbalanced data classification \cite{fernandez2018smote}, motivating numerous extensions to the original approach. A common theme, shared by a majority of these SMOTE-derived methods, is a mechanism of focusing the oversampling in certain regions. Examples of such approaches include Borderline-SMOTE \cite{han2005borderline} which, as the name indicates, focuses resampling near the borderline region; Safe-Level-SMOTE \cite{Bunkhumpornpat:2009}, in which observations are generated near the computed safe regions; ADASYN \cite{he2008adasyn}, which produces higher quantity of synthetic observations around difficult observations; and MWMOTE \cite{barua2012mwmote}, also focusing on the difficult observations. More recently, density-based approaches utilizing radial basis functions, such as Radial-Based Oversampling \cite{koziarski2019radial} and Sampling With the Majority Class \cite{bellinger2019framework}, also display the behavior of modifying the underlying class distribution. All of the aforementioned resampling strategies are based on different, often contradictory, ideas on where the resampling process should be focused. While all of them have their own niches of outperformance, out of necessity they are specialized, and it is often not clear which method, if any, is preferred in a general case.

Instead of using an ad-hoc strategy of boosting specific regions of data space, in this paper we propose taking the approach of preserving the original shape of underlying class distribution. Specifically, we achieve that by treating the generated synthetic observations as optimization parameters, which are positioned to minimize the difference between the potential of the original and resampled observations, with a regularization constraint added to prevent overfitting during oversampling.

\subsection{Potential resemblance loss}

We base our approach on the concept of class potential. Potential functions were previously used in the context of imbalanced data resampling in \cite{koziarski2019radial}. Given a collection of observations $\mathcal{X}$, and a spread of a single radial-basis function $\gamma$, potential in a given point in space $x$ can be defined as 
\begin{equation}
    \Phi(\mathcal{X}, x, \gamma) = \sum_{i=1}^{\mid \mathcal{X} \mid}{e^{-\left(\frac{\lVert \mathcal{X}_i - x \rVert_2}{\gamma}\right)^{2}}}.
    \label{eq:potential}
\end{equation}

Intuitively, potential can be viewed as a measure of cumulative proximity of $x$ to $X$, with higher value of potential indicating that more observations from $X$ lie in a close proximity to $x$. Of particular interest to our discussion are majority and minority class potential, computed on, respectively, a collection of majority class observations $\mathcal{X}_{maj}$ and minority class observations $\mathcal{X}_{min}$. Class potential can be viewed as a measure of density of observations from that class.

The values of potential function are not, however, bound to any specific range, making it difficult to compare the relative shape of potential computed with respect to two different collections of observations. To mitigate this issue we propose a normalized potential function $\Psi$. This function computes the potential for $k$ anchor points $\mathcal{A}$, and returns a vector of $k$ normalized potentials. More formally, we define the normalized potential function as
\begin{equation}
    \Psi(\mathcal{X}, \mathcal{A}, \gamma) = \frac{1}{\sum_{i=1}^{k}{\Phi(\mathcal{A}_i, \mathcal{X}, \gamma)}} %\cdot 
    \left[
    \begin{gathered}
    \Phi(\mathcal{A}_1, \mathcal{X}, \gamma) \\
    \Phi(\mathcal{A}_2, \mathcal{X}, \gamma) \\
    ... \\
    \Phi(\mathcal{A}_{k}, \mathcal{X}, \gamma)
    \end{gathered}
    \right].
    \label{eq:normalized-potential}
\end{equation}

Due to the non-negativity of $\Phi$, the values of $\Psi$ are also non-negative and range from 0 to 1. The normalized potential function describes the relative density of observations in any given anchor point $\mathcal{A}_i$. This property makes it possible to directly compare the outputs of $\Psi$ computed with respect to two different collections of observations, even if the collections differ in size, as will be the case during resampling.

Finally, based on the concept of normalized potential we define the potential resemblance loss. Given a collection of original observations $\mathcal{X}$, $k$ anchor points $\mathcal{A}$, collection of prototypes, that is generated observations the position of which we wish to optimize, $\mathcal{P}$, their starting positions $\mathcal{P}^0$, radial basis function spread $\gamma$, and regularization coefficient $\lambda$, we define the potential resemblance loss as
\begin{equation}
    \begin{aligned}
    \mathcal{L}(\mathcal{X}, \mathcal{A}, \mathcal{P}, \mathcal{P}^0, \gamma, \lambda) = & \sum_{i=1}^{k}{(\Psi(\mathcal{X}, \mathcal{A}, \gamma)_i - \Psi(\mathcal{P}, \mathcal{A}, \gamma)_i)^2} \\ 
    & + \lambda  \sum_{i=1}^{\mid \mathcal{P} \mid}{e^{-\left(\frac{\lVert \mathcal{P}_i - \mathcal{P}^0_i \rVert_2}{\gamma}\right)^{2}}}
    \end{aligned}.
    \label{eq:loss}
\end{equation}

The left-side term of the equation is a mean squared error between the normalized potential computed with respect to $\mathcal{X}$ and $\mathcal{P}$, and as such measures the difference in relative shape of the potential produced by these two collection of observations. The right-side term is a regularization term that measures the displacement of prototypes $\mathcal{P}$ from their starting positions. If $\mathcal{P}^0$ is created by a random sampling of $\mathcal{X}$, as  will be the case in the proposed approach, this term prevents the algorithm minimizing $\mathcal{L}$ from degenerating to random oversampling, since prototypes $\mathcal{P}$ will be displaced from their original positions. It is worth noting that even though in principle we would like to penalize the placement of $\mathcal{P}$ close to any of the observations from $\mathcal{X}$, in our experiments considering only the starting positions $\mathcal{P}^0$ was sufficient to prevent the overfitting, at the same time being more computationally efficient.

\subsection{Algorithm}

Being equipped with a potential resemblance loss $\mathcal{L}$, we can then formulate the problem of imbalanced data resampling as the optimization of prototype point positions $\mathcal{P}$ with respect to $\mathcal{L}$. While the proposed approach is motivated from the point of view of oversampling, it is also easily applicable to the undersampling, as the same principle of preserving the original class density can be applied in both cases. The main difference between the two is that while during the oversampling we can preserve the original minority observations and use prototypes $\mathcal{P}$ as a collection of additional, synthetic observations, during the undersampling we will instead replace the original majority observations with a smaller collection of prototypes preserving the original class potential.

We present the pseudocode of the proposed Potential Anchoring (PA) algorithm in Algorithm~\ref{algorithm:pa}. The method combines over- and undersampling up to the point of achieving balanced class distribution, with the ratio of imbalance eliminated with either over- or undersampling treated as a parameter. First, $k$ anchor points, with respect to which normalized potential will be calculated, are generated via clustering of the collection of original observations $\mathcal{X}$. Second, the prototypes are initialized by randomly sampling with replacement from the collection of observations of a given class. Importantly, small random jitter is afterwards introduced to break the symmetry during the optimization. Finally, the prototypes are then optimized with respect to the potential resemblance loss function $\mathcal{L}$, separately for the majority and the minority class. This loss function is penalized with a regularization coefficient $\lambda$ in the case of oversampling. Throughout the conducted experiments we used $k$-means clustering to generate the anchor points. We also leveraged the differentiability of $\mathcal{L}$ and conducted the optimization using Adam optimizer \cite{kingma2014adam}.

\begin{algorithm*}[!htb]
	\caption{Potential Anchoring}
	\textbf{Input:} collection of original observations $\mathcal{X}$ divided into majority $\mathcal{X}_{maj}$ and minority $\mathcal{X}_{min}$ observations \\
	\textbf{Parameters:} $ratio$ of imbalance eliminated with oversampling, number of anchor points $k$, number of $iterations$, radial basis function spread $\gamma$, oversampling regularization coefficient $\lambda$, learning rate $\alpha$, random jitter used for initialization $\epsilon$ \\
    \textbf{Output:} collection of resampled observations $\mathcal{X}'$
		
	\label{algorithm:pa}	
	\vspace{-0.5\baselineskip}
	
	\hrulefill
	\begin{algorithmic}[1]
		\STATE \textbf{function} PA($\mathcal{X}_{maj}$, $\mathcal{X}_{min}$, $ratio$, $k$, $iterations$, $\gamma$, $\lambda$, $\alpha$, $\epsilon$):
		\STATE $\mathcal{A} \gets k$ anchor points obtained by clustering $\mathcal{X}$
		\STATE $n_{PAO} \gets ratio \cdot \left( \lvert \mathcal{X}_{maj} \rvert - \lvert \mathcal{X}_{min} \rvert \right)$
		\STATE $n_{PAU} \gets$ $\lvert \mathcal{X}_{maj} \rvert - \left(1 - ratio\right) \cdot \left( \lvert \mathcal{X}_{maj} \rvert - \lvert \mathcal{X}_{min} \rvert \right)$
		\STATE $\mathcal{P}_{min}^0 \gets n_{PAO}$ prototypes randomly selected with replacement from $\mathcal{X}_{min}$
		\STATE $\mathcal{P}_{maj}^0 \gets n_{PAU}$ prototypes randomly selected with replacement from $\mathcal{X}_{maj}$
		\STATE $\mathcal{P}_{min}, \mathcal{P}_{maj} \gets \mathcal{P}_{min}^0, \mathcal{P}_{maj}^0 $ with added random jitter $\epsilon$
		\FOR{i in 1..$iterations$}
		    \STATE perform optimization step on $\mathcal{P}_{min}$ w.r.t. $\mathcal{L}(\mathcal{X}_{min}, \mathcal{A}, \mathcal{P}_{min}, \mathcal{P}_{min}^0, \gamma, \lambda)$ using learning rate $\alpha$
		\ENDFOR
		\FOR{i in 1..$iterations$}
		    \STATE perform optimization step on $\mathcal{P}_{maj}$ w.r.t. $\mathcal{L}(\mathcal{X}_{maj}, \mathcal{A}, \mathcal{P}_{maj}, \mathcal{P}_{maj}^0, \gamma, 0)$ using learning rate $\alpha$
		\ENDFOR
		\STATE $\mathcal{X}' \gets \mathcal{X}_{min} \cup \mathcal{P}_{min} \cup \mathcal{P}_{maj}$
		\STATE \textbf{return} $\mathcal{X}'$
	\end{algorithmic}
\end{algorithm*}

\begin{figure}[!htb]
\centering
\begin{subfigure}[b]{0.22\textwidth}
  \includegraphics[width=\textwidth]{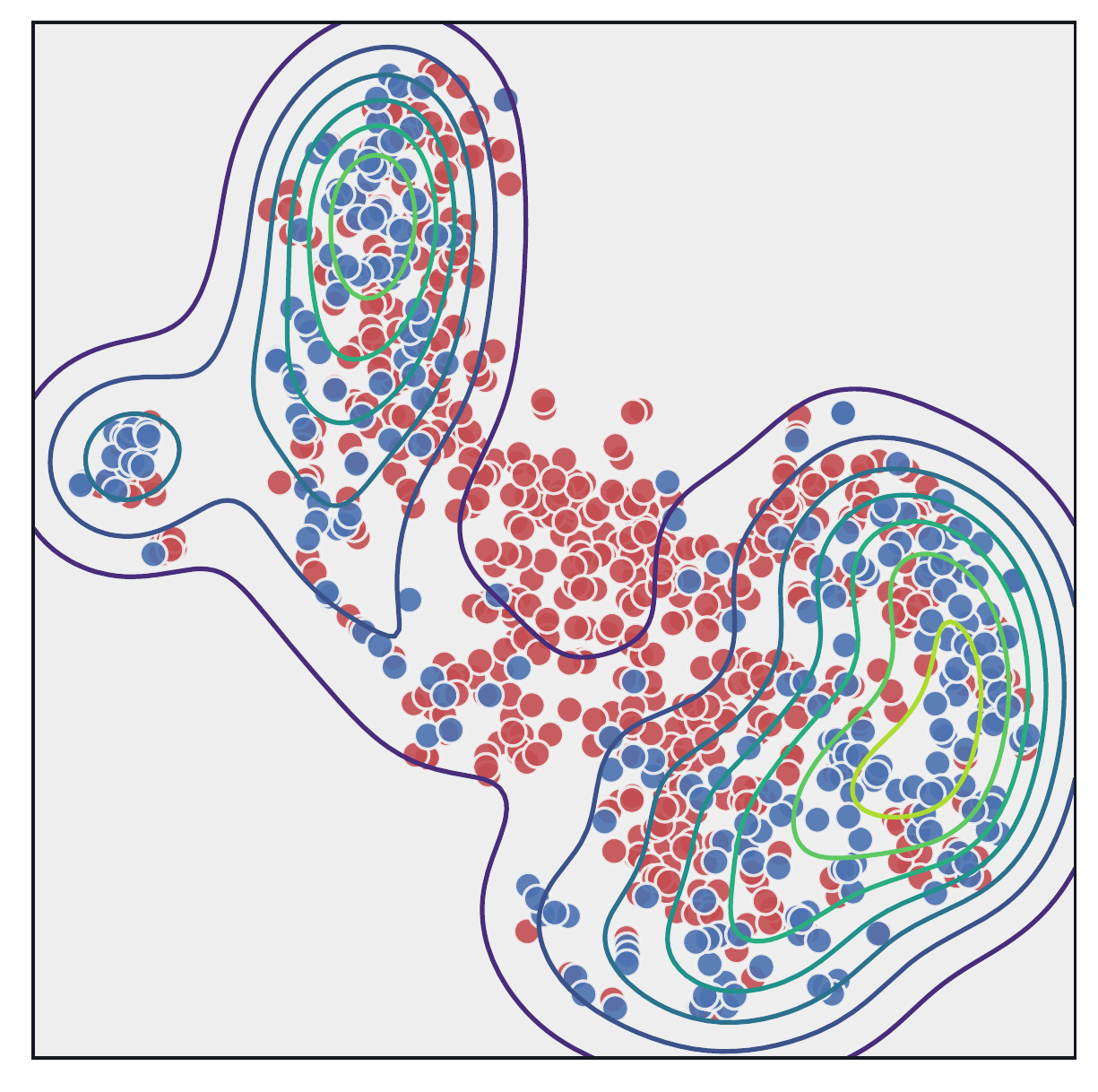}
  \caption{minority class potential}
\end{subfigure}
~
\begin{subfigure}[b]{0.22\textwidth}
  \includegraphics[width=\textwidth]{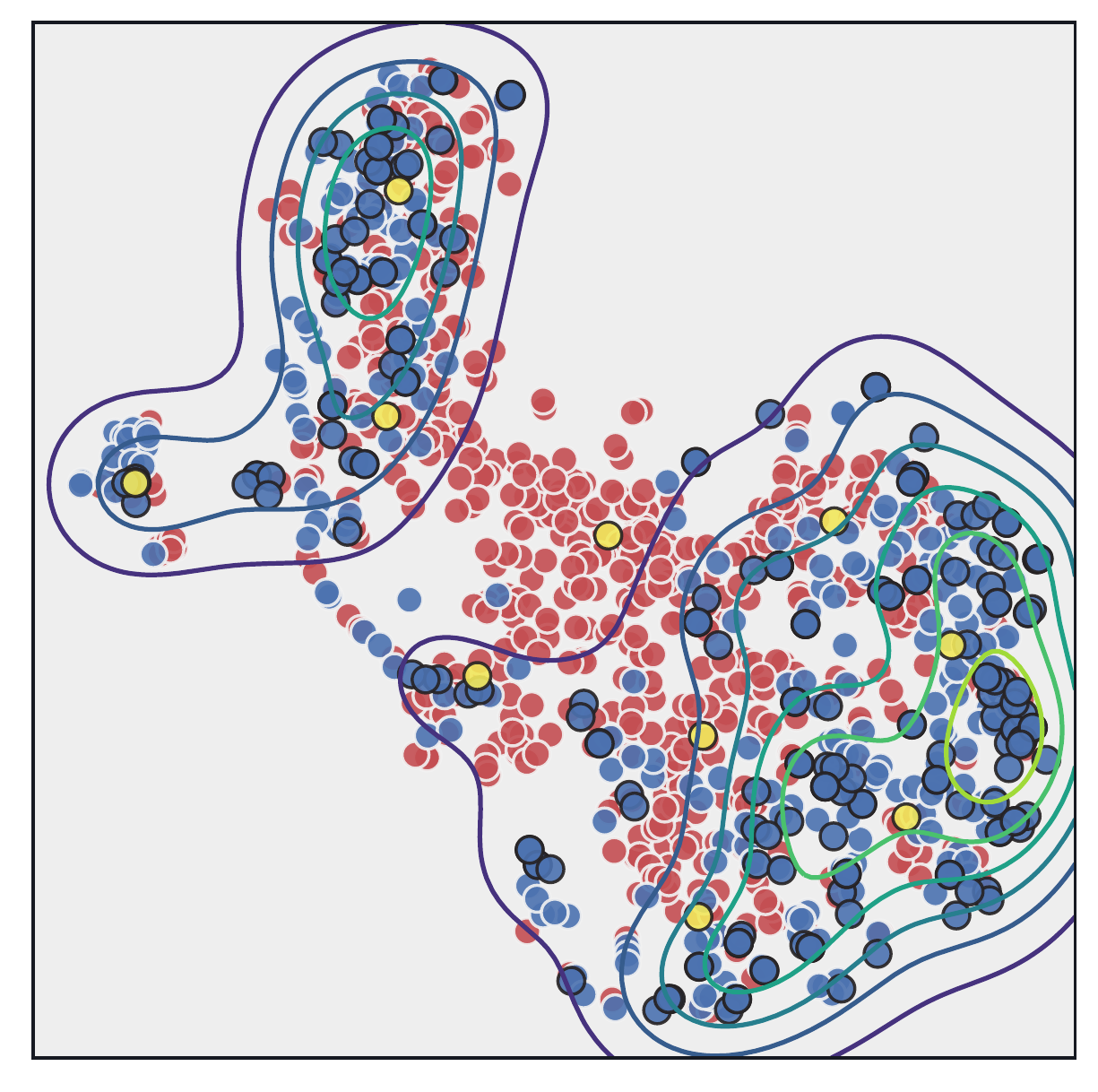}
  \caption{PAO oversampling}
\end{subfigure}

\begin{subfigure}[b]{0.22\textwidth}
  \includegraphics[width=\textwidth]{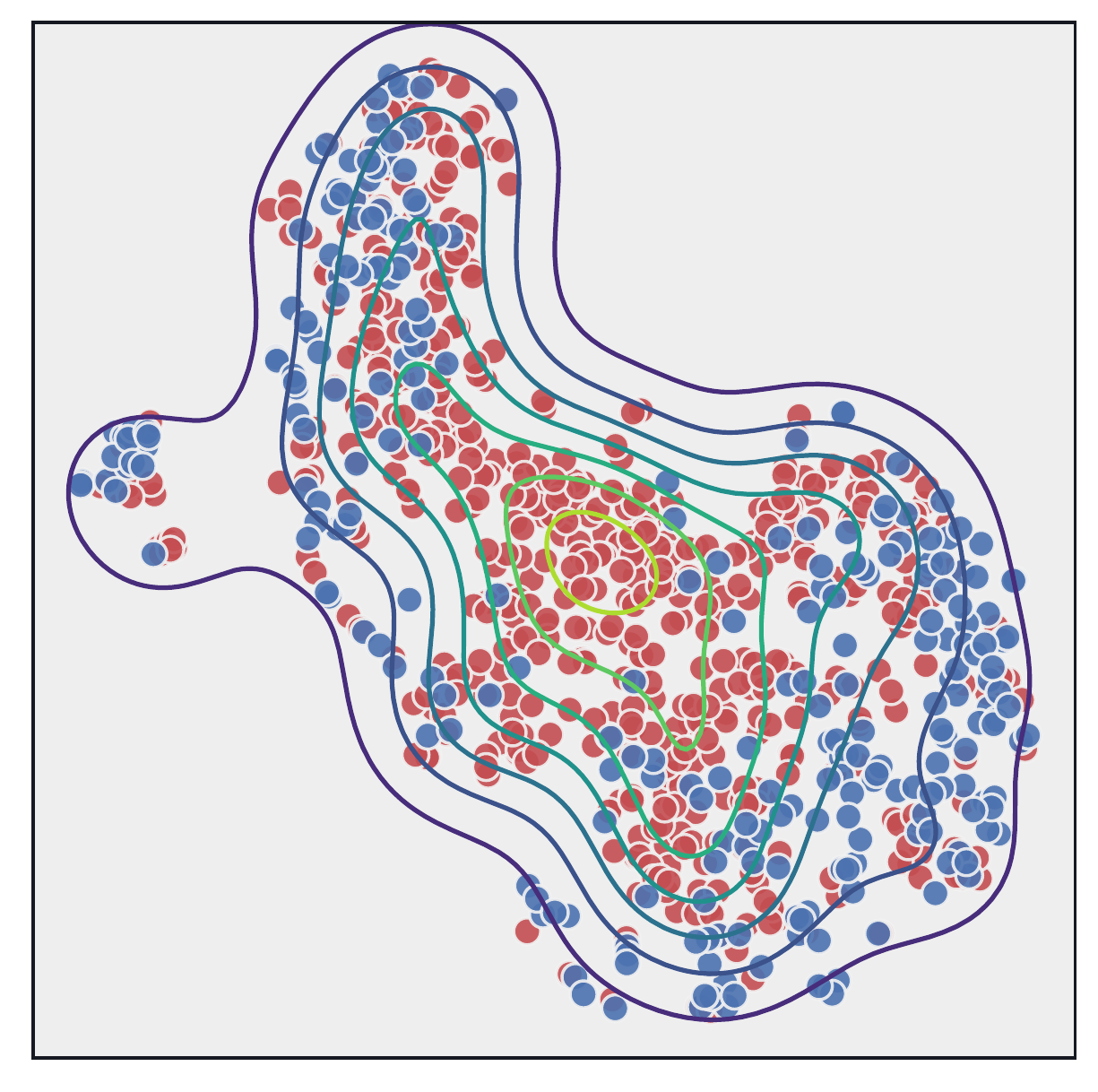}
  \caption{majority class potential}
\end{subfigure}
~
\begin{subfigure}[b]{0.22\textwidth}
  \includegraphics[width=\textwidth]{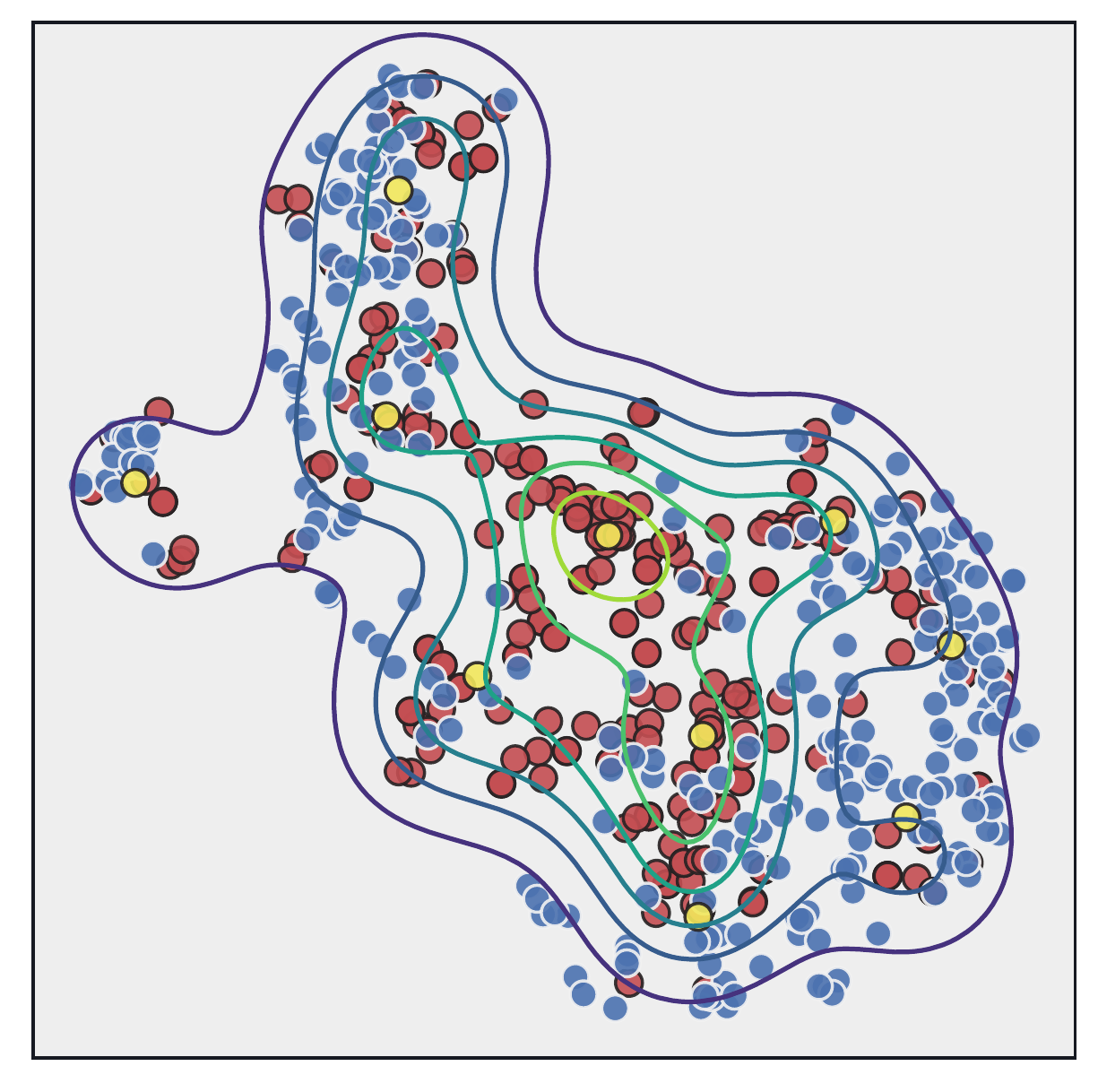}
  \caption{PAU undersampling}
\end{subfigure}
\caption{Example of the original minority and majority class potentials compared with the potential of synthetic observations generated by PAO and PAU. Anchor points denoted with a yellow color.}
\label{fig:example-pao-pau}
\end{figure}

\begin{figure*}[!htb]
\centering
\begin{subfigure}[b]{0.22\textwidth}
  \includegraphics[width=\textwidth]{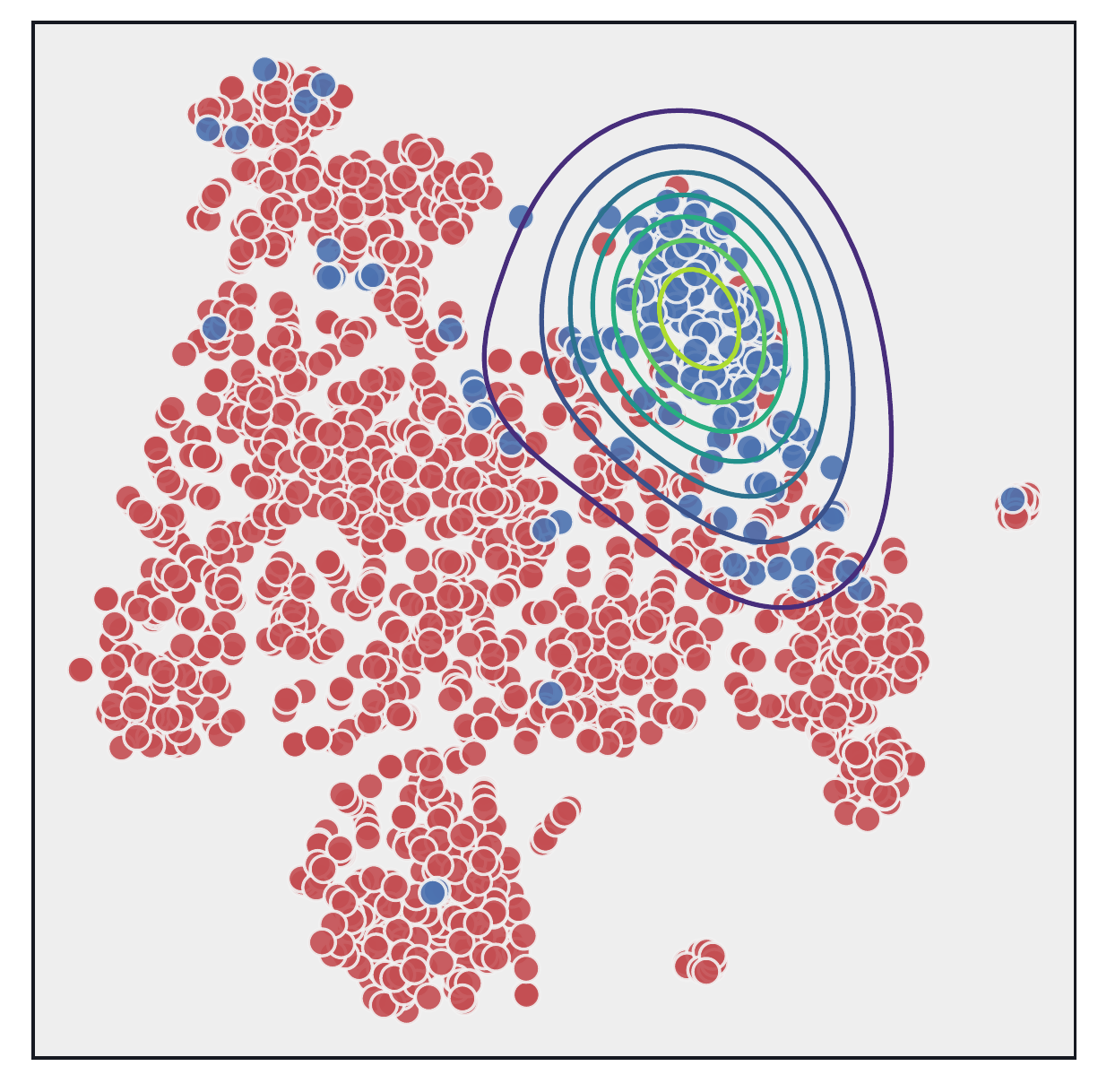}
  \caption{original dataset}
\end{subfigure}
~
\begin{subfigure}[b]{0.22\textwidth}
  \includegraphics[width=\textwidth]{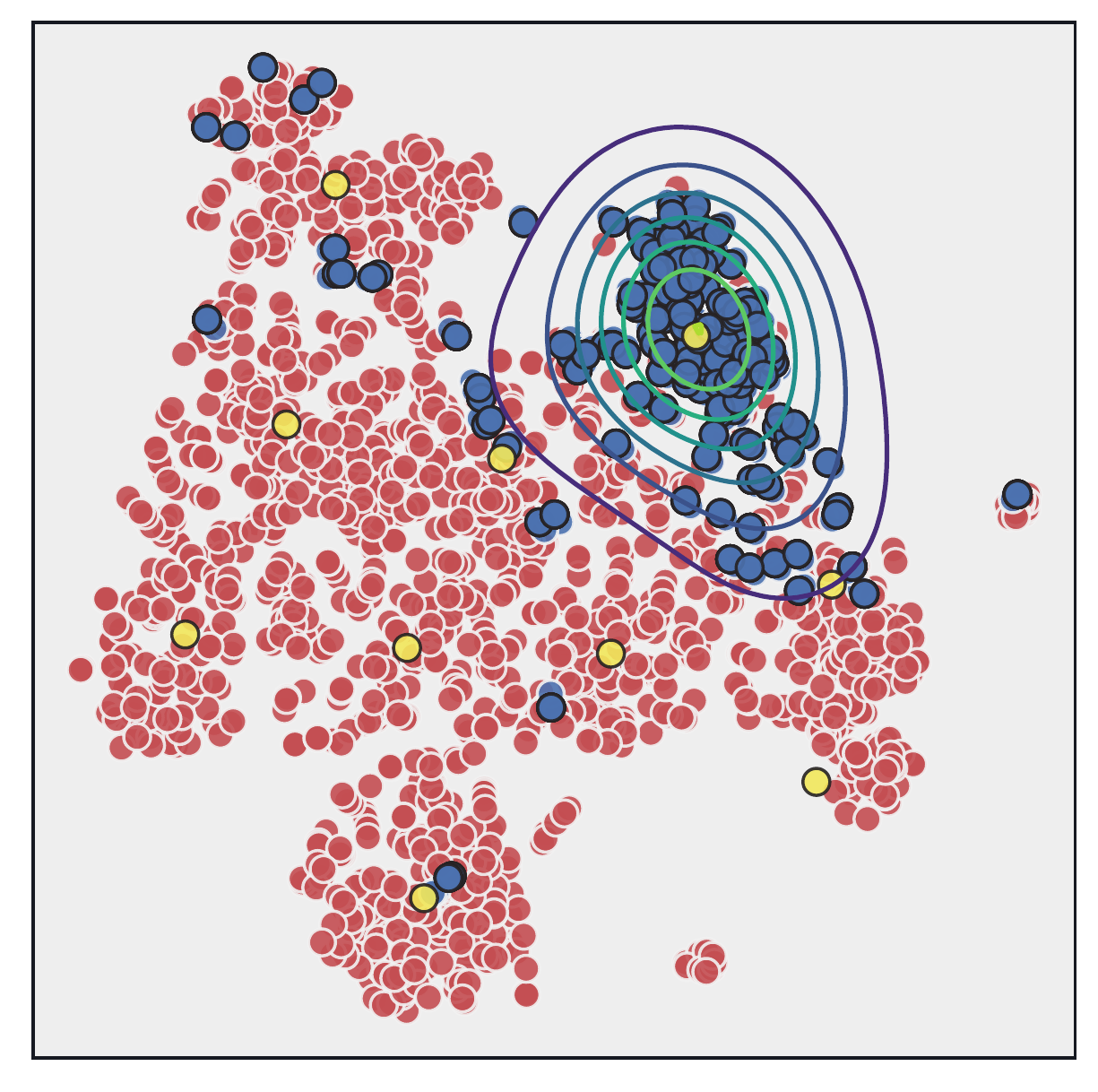}
  \caption{$\lambda = 0.0$}
\end{subfigure}
~
\begin{subfigure}[b]{0.22\textwidth}
  \includegraphics[width=\textwidth]{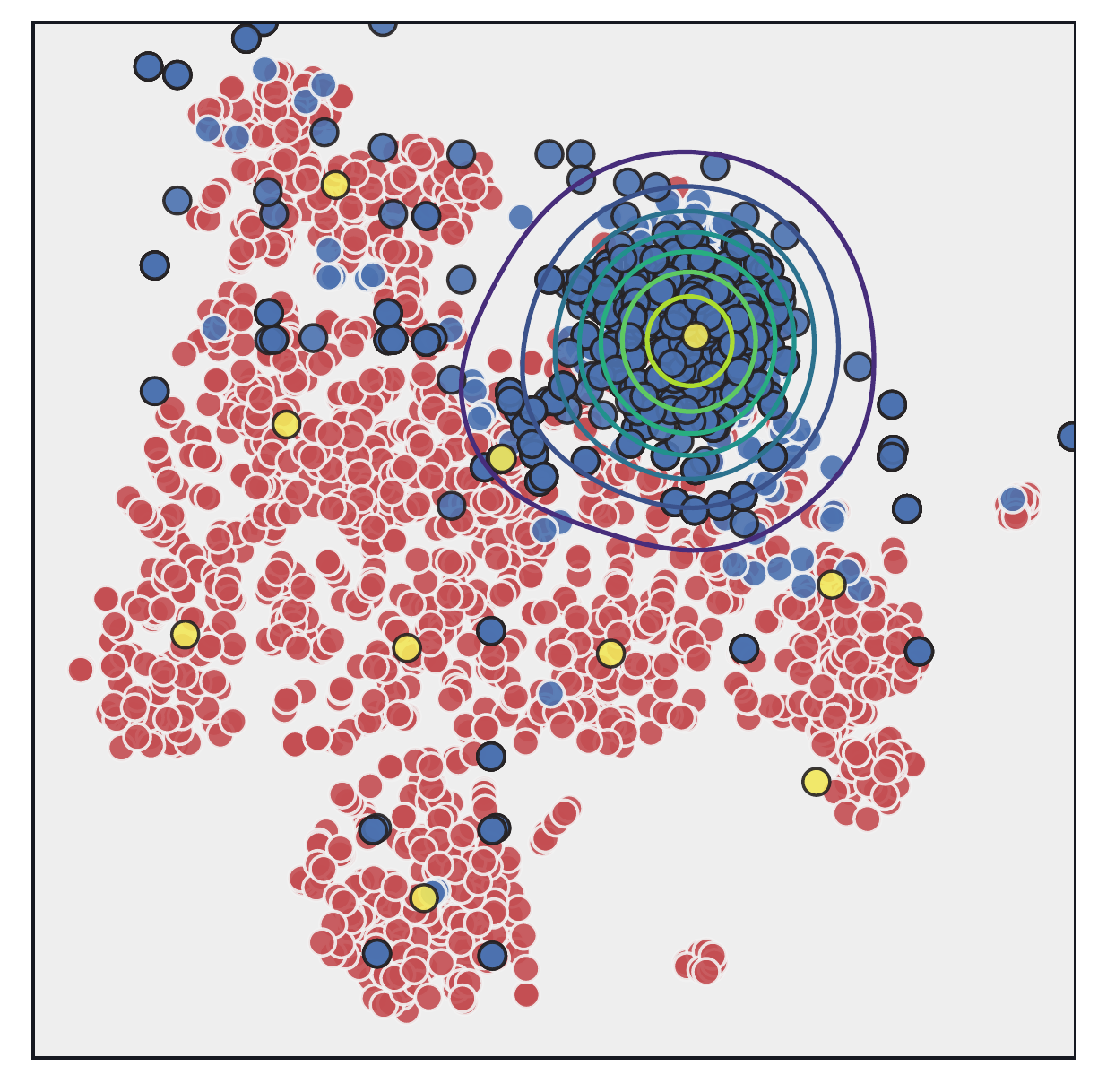}
  \caption{$\lambda = 0.001$}
\end{subfigure}
~
\begin{subfigure}[b]{0.22\textwidth}
  \includegraphics[width=\textwidth]{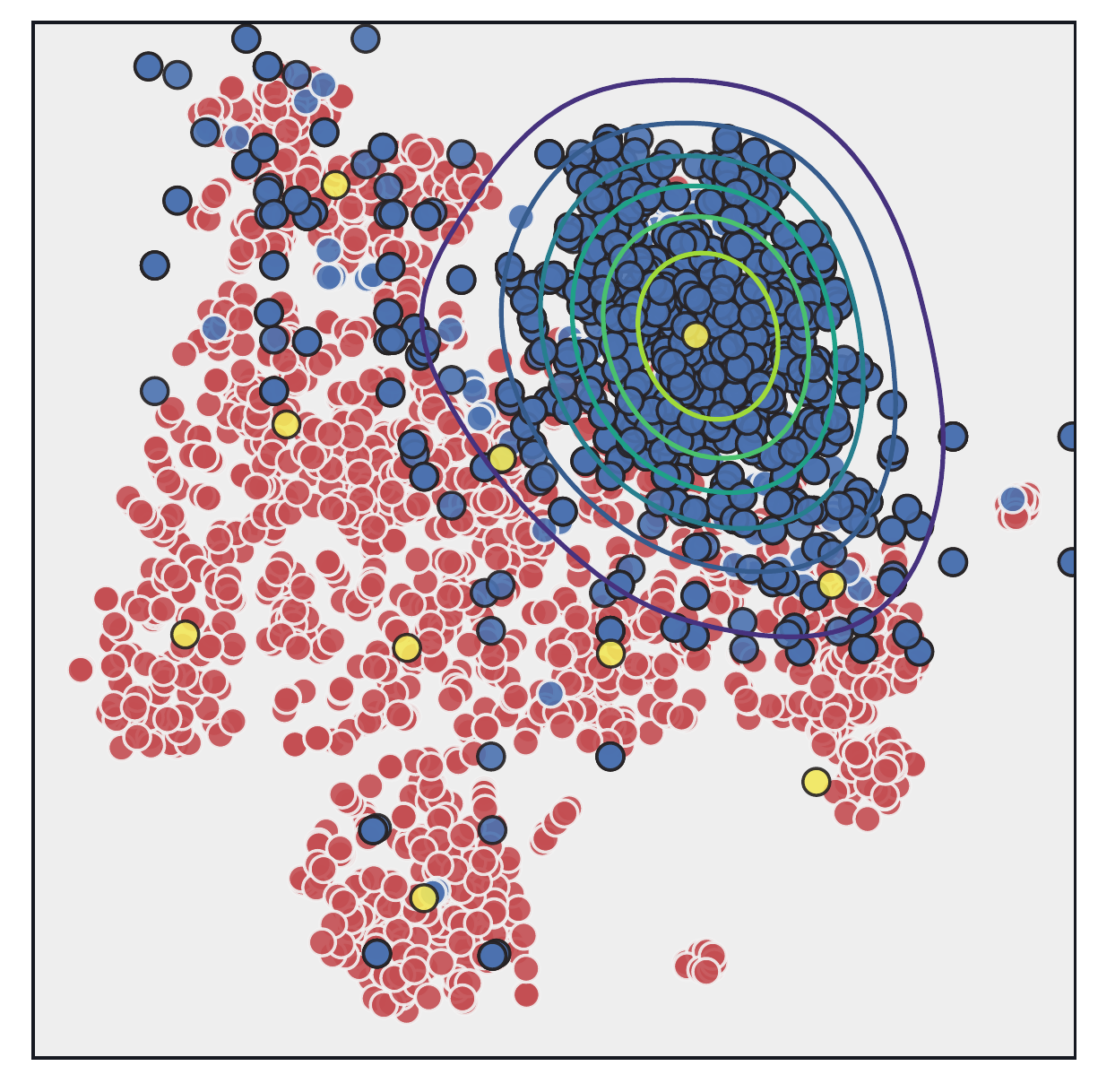}
  \caption{$\lambda = 10.0$}
\end{subfigure}
\caption{Visualization of the impact of $\lambda$ parameter on the behavior of PAO. Higher values of $\lambda$ lead to lower similarity between the original and generated potential shape, and higher spread of synthesized observations.}
\label{fig:example-lambda}
\end{figure*}

\begin{figure*}[!htb]
\centering
\begin{subfigure}[b]{0.22\textwidth}
  \includegraphics[width=\textwidth]{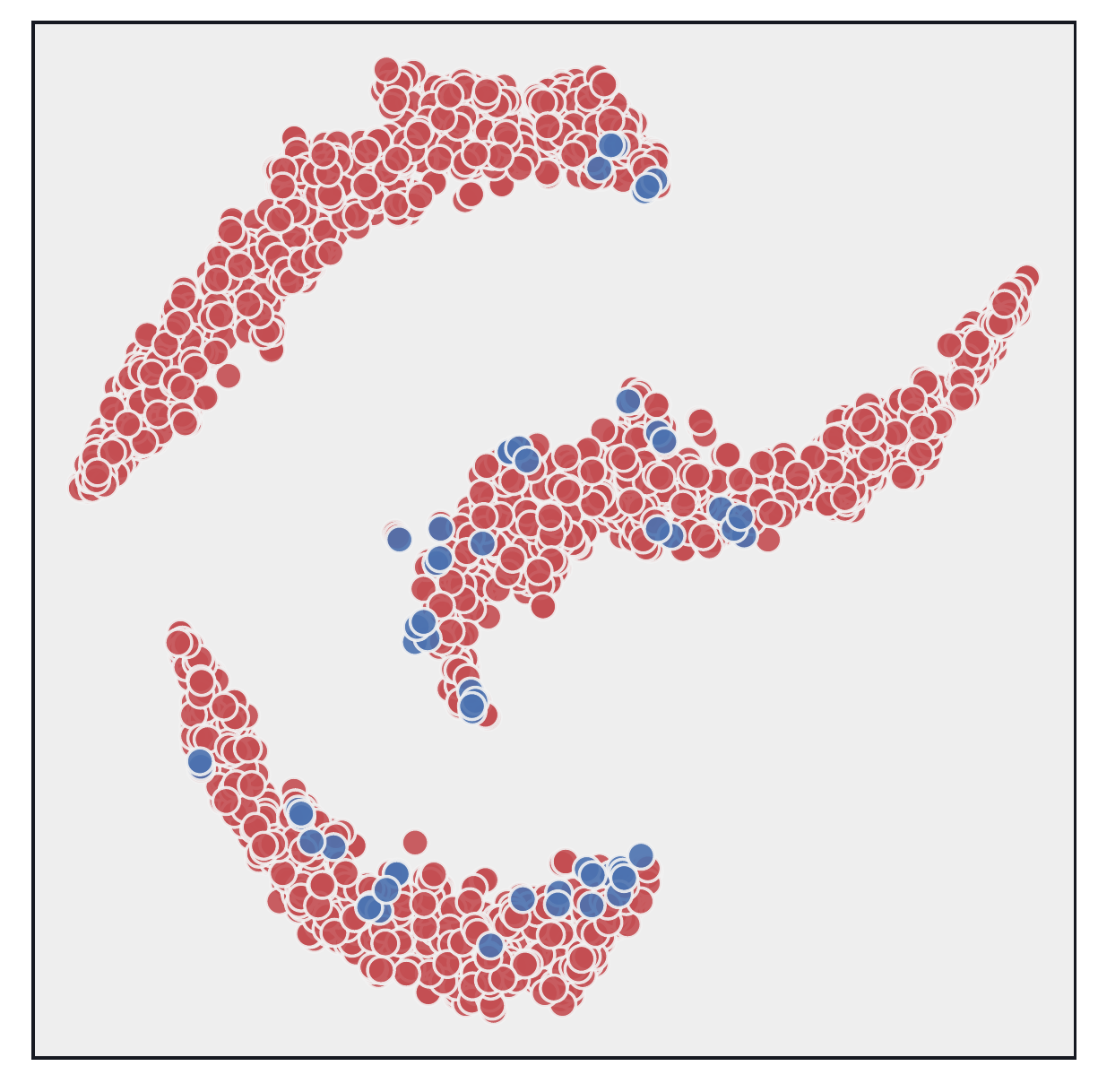}
  \includegraphics[width=\textwidth]{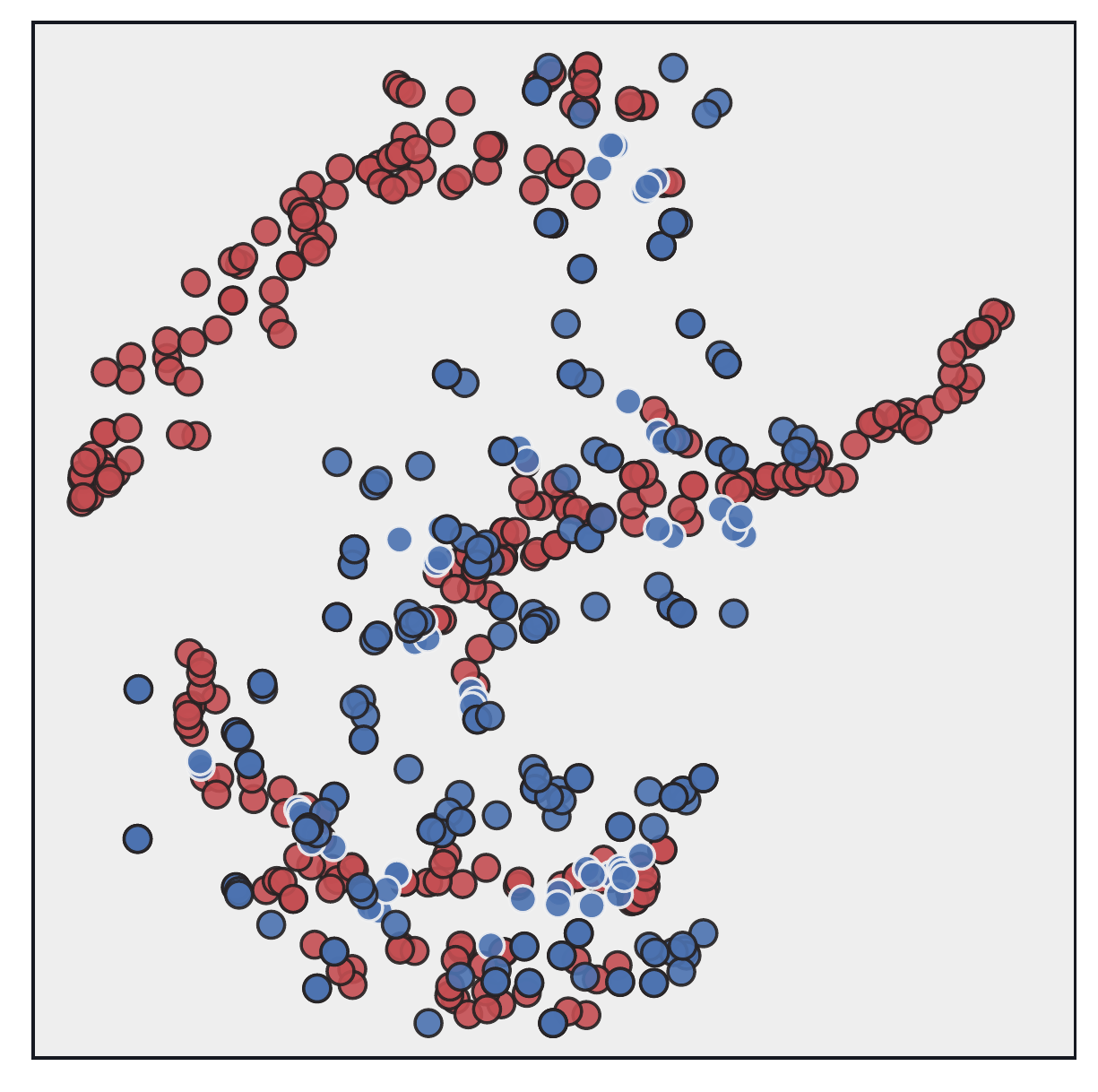}
\end{subfigure}
~
\begin{subfigure}[b]{0.22\textwidth}
  \includegraphics[width=\textwidth]{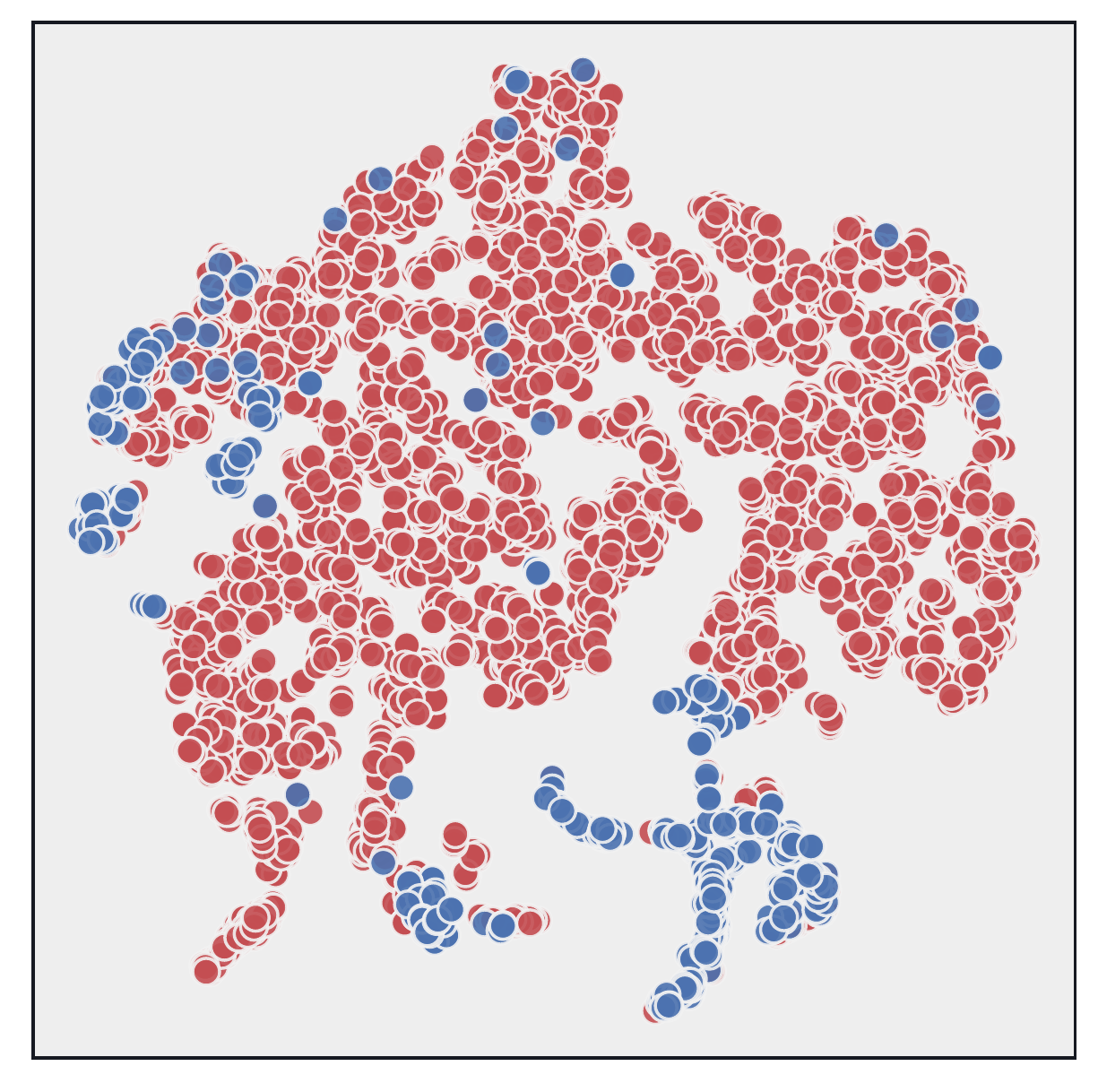}
  \includegraphics[width=\textwidth]{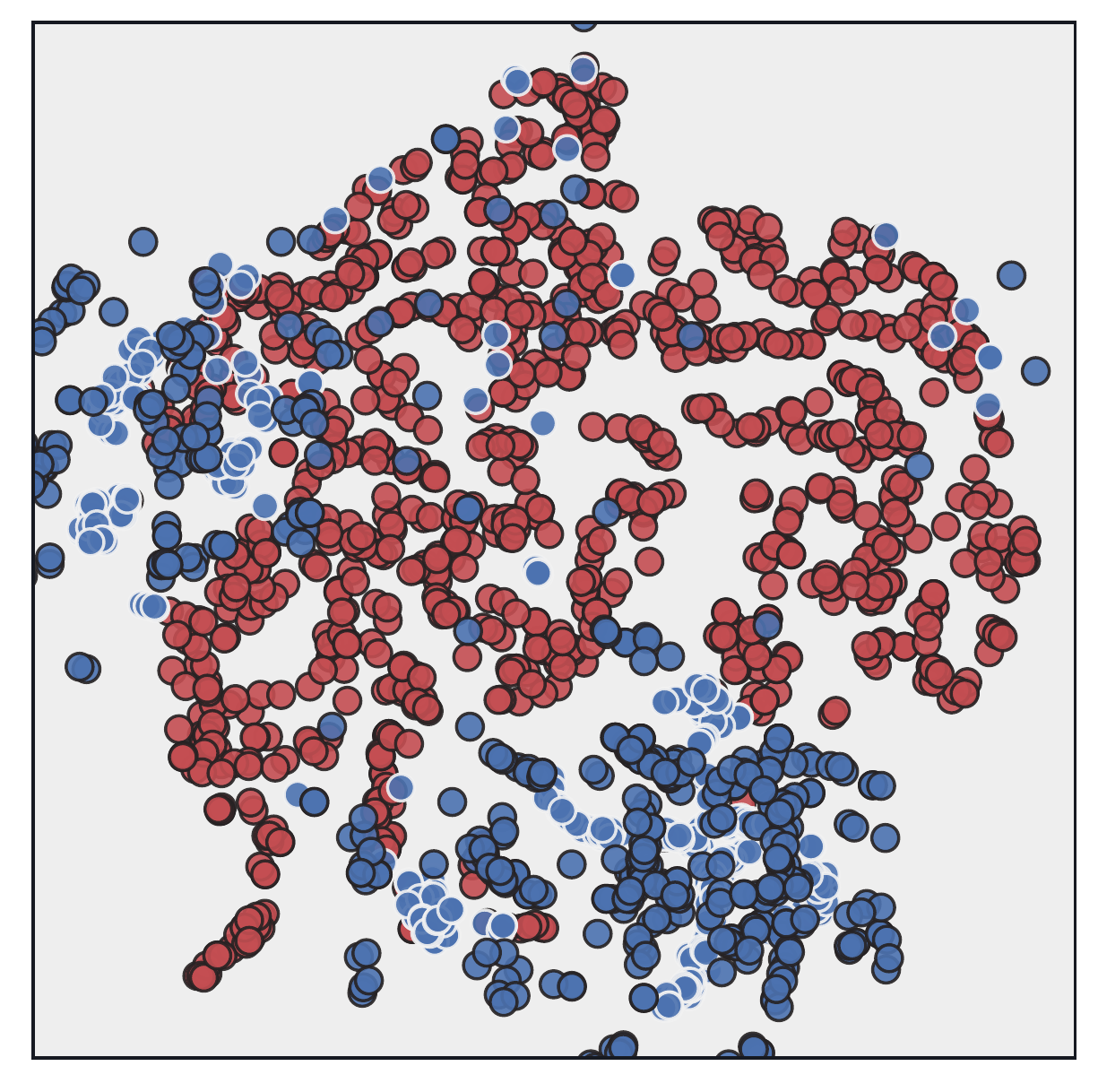}
\end{subfigure}
~
\begin{subfigure}[b]{0.22\textwidth}
  \includegraphics[width=\textwidth]{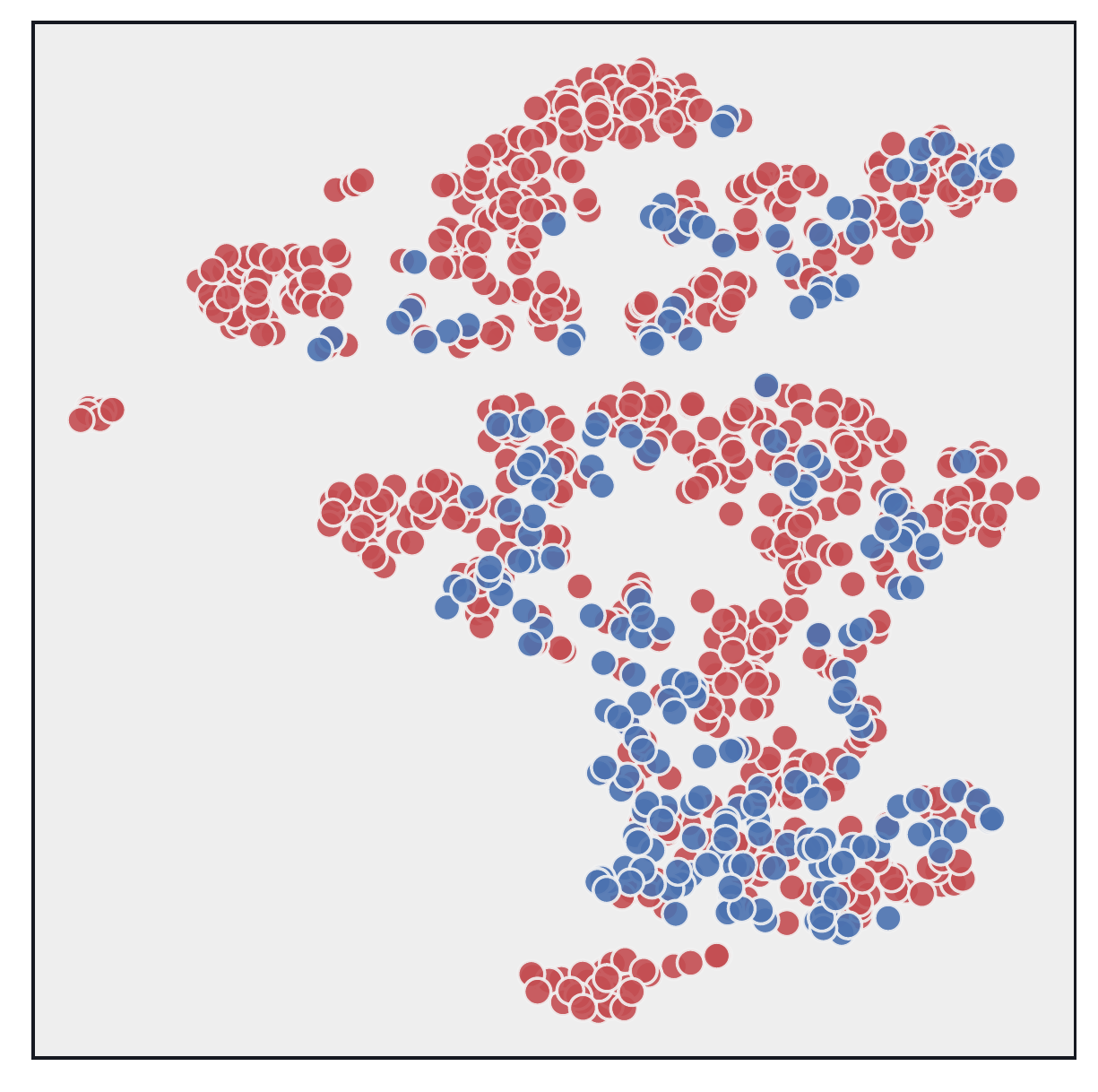}
  \includegraphics[width=\textwidth]{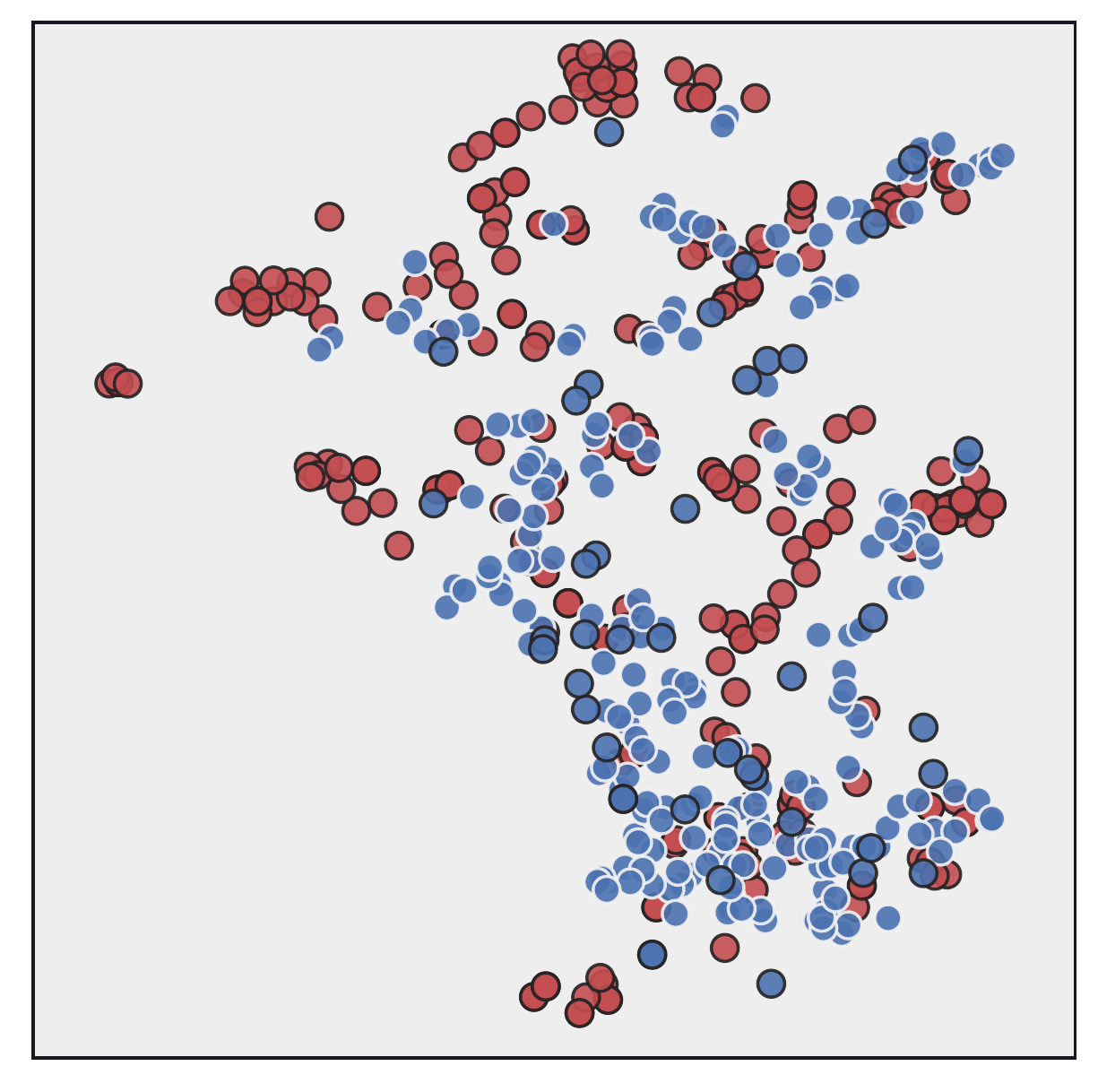}
\end{subfigure}
~
\begin{subfigure}[b]{0.22\textwidth}
  \includegraphics[width=\textwidth]{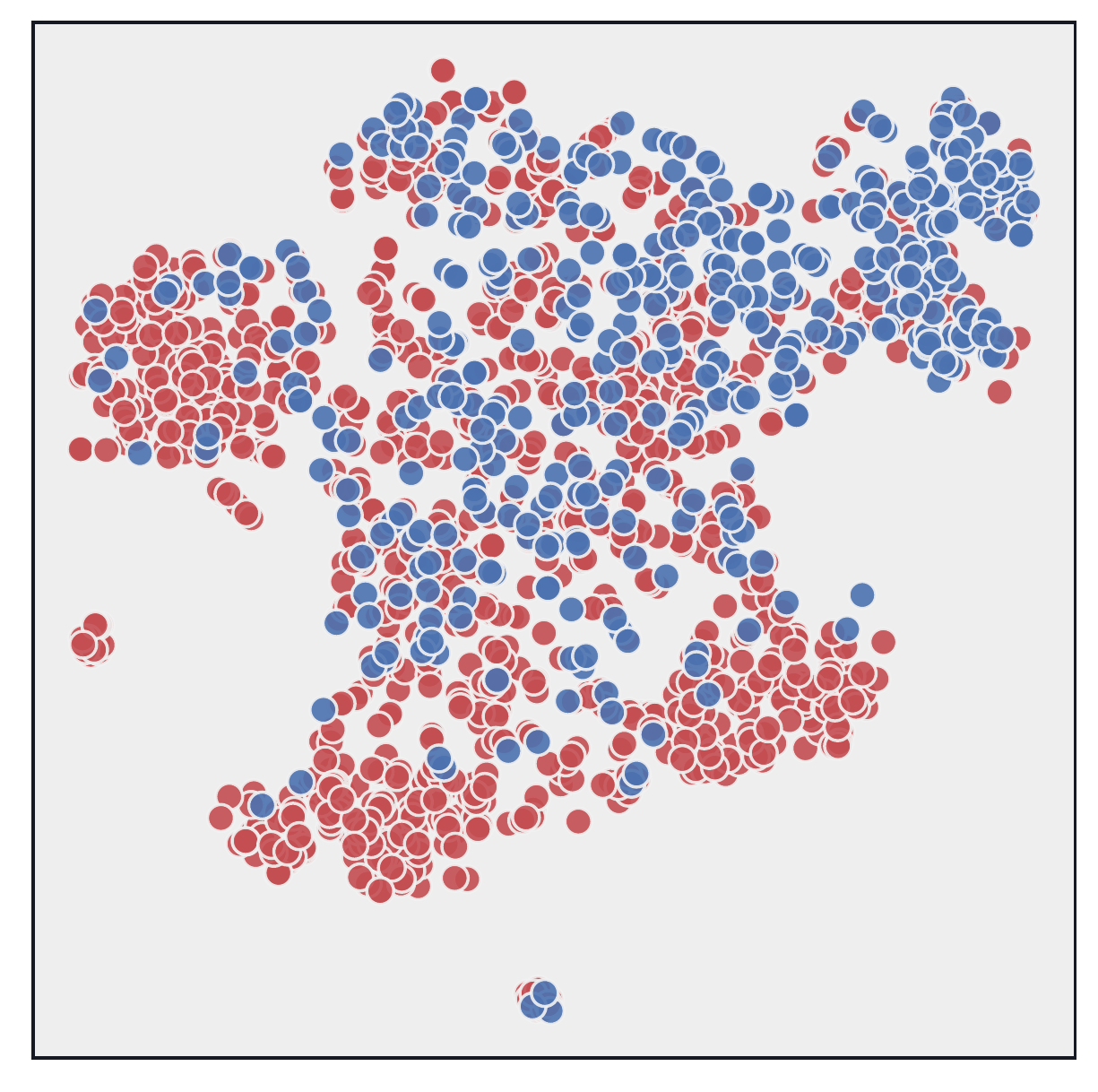}
  \includegraphics[width=\textwidth]{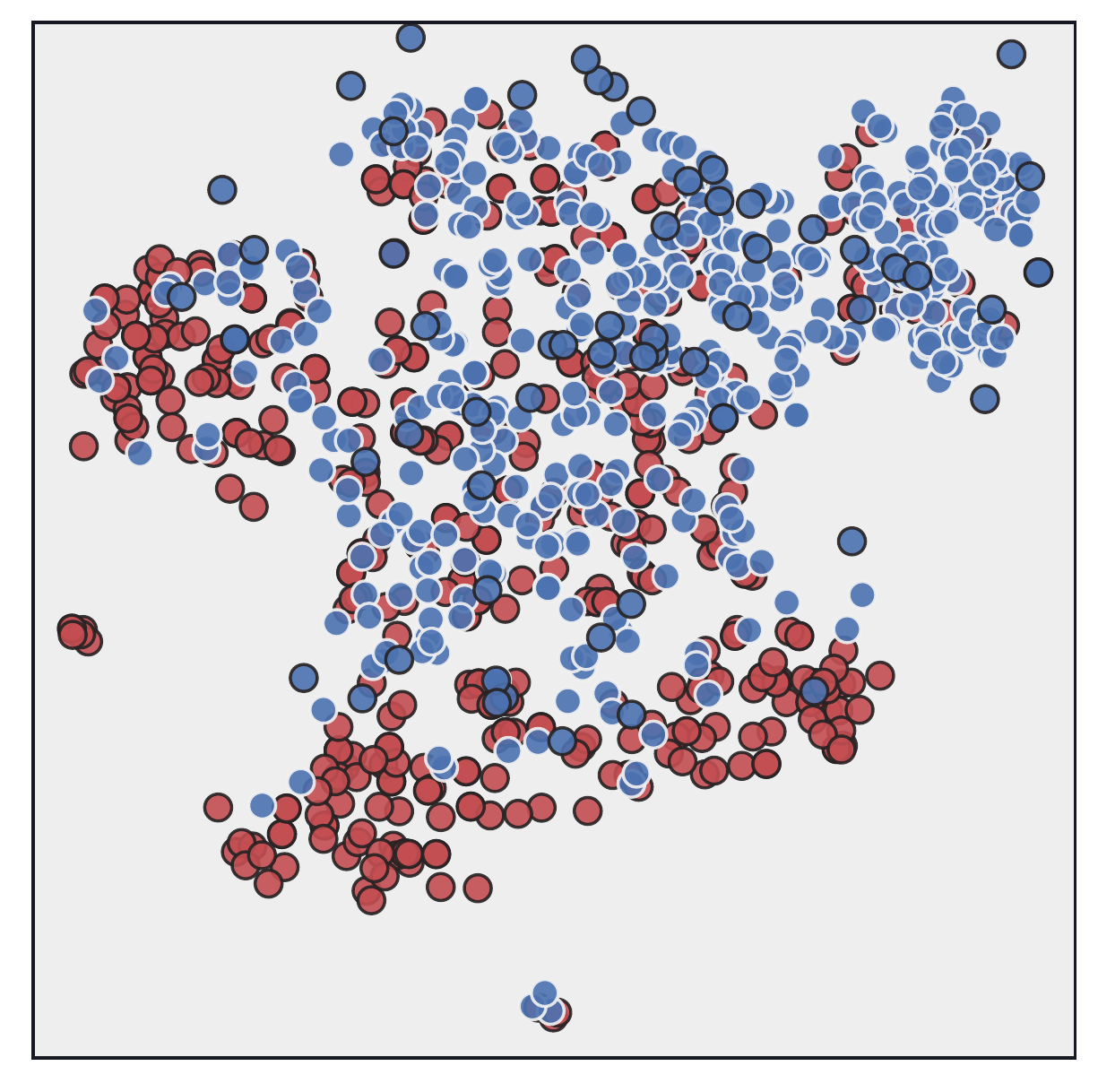}
\end{subfigure}
\caption{Example of PA resampling applied on datasets with different imbalance characteristics. Top row: original datasets, bottom row: datasets after resampling.}
\label{fig:example-pa}
\end{figure*}

A particular case of PA involves eliminating the imbalance solely by either oversampling (PAO) or undersampling (PAU). We illustrate the concept of class potential in both cases in Figure~\ref{fig:example-pao-pau}. As can be seen, PA generates synthetic observations, the potential of which resembles the original despite the fact that it is being anchored in a small number of points. Secondly, we illustrate the impact of regularization coefficient $\lambda$ on the behavior of PAO in Figure~\ref{fig:example-lambda}. As can be seen, higher values of $\lambda$ lead to lower similarity between the original and generated potential shape, and higher spread of synthesized observations. Disabling regularization leads to minimal translation of the prototypes, and behavior closely resembling random oversampling. Finally, we illustrate the behavior of PA, with both over- and undersampling used, and regularization enabled, in Figure~\ref{fig:example-pa}.

\section{Data difficulty index}
\label{sec:data-difficulty-index}

An important aspect of experimental studies involving imbalanced data resampling algorithms, or machine learning algorithms in general, is identification of their areas of applicability. According to the "no free lunch" theorem \cite{wolpert1997no}, we should not expect any single algorithm to achieve an optimal performance in every considered problem. Instead, we can identify the conditions under which the considered algorithm tends to outperform the reference methods. This serves two goals. First, to provide a rule of thumb for a practitioner deciding whether the usage of our algorithm is sensible. Second, to guide the future research, by focusing on the detected strengths and weaknesses of the proposed method during further development.

One of the most important factors affecting the performance during imbalanced data classification is the complexity of the considered datasets. Recent studies \cite{stefanowski2016dealing,Fernandez:2018} recognize that data imbalance, by itself, does not have to pose a challenge for learning algorithms. Instead, it exacerbates the negative impact of other data difficulty factors, such as small sample size, presence of disjoint and overlapping data distributions, and presence of outliers and noisy observations. It is therefore beneficial to evaluate the impact of said factors on the performance of the proposed algorithms.

A recent methodology proposed by Koziarski \cite{koziarski2020radial} examined relation between the proportion of difficult observations and relative performance observed for a given algorithm. This methodology utilized the categorization introduced by Napierała and Stefanowski \cite{napierala2016types}, in which minority observations are assigned one of four categories based on their nearest neighborhood. Specifically, the category is assigned based on the number of nearest neighbors from the same class: \textit{safe} in case of 4 to 5 neighbors from the same class, \textit{borderline} in case of 2 to 3 neighbors, \textit{rare} in case of 1 neighbor, and \textit{outlier} when there are no neighbors from the same class. The proportion of observations from any given category was afterwards correlated across multiple datasets with the rank achieved by the considered algorithm on a given dataset. Afterwards, conclusions could have been drawn about either increase or decrease in the relative performance, depending on the proportion of observations from a given category. This analysis was conducted separately for each observation category.

One shortcoming of this approach was caused by the fact that not every dataset consists of observations from each category, which could have a confounding effect on the results of the analysis. To give an example, we can consider two cases of datasets, the first consisting entirely of rare minority observations, and the second entirely of outliers. While both datasets can clearly be classified as difficult, the first one will not be treated as such during the examination of outlier percentage (since there are none), and the second during the examination of percentage of rare observations. This independence of analyses conducted with respect to different observation categories can have an unwanted effect on the results.

To address this issue we propose extending the concept of aforementioned categorization into a data difficulty index (DI), a function that measures the average proportion of majority neighbors for each minority observation. More formally, let us define the collection of all observations by $\mathcal{X}$, the collection of $n_{min}$ minority observations by $\mathcal{X}_{min}$, $i$-th minority observation as $\mathcal{X}^{(i)}_{min}$, and a function returning $j$-th nearest neighbor of observation $x$ from the collection of $\mathcal{X}$ as $f_{NN}(x, \mathcal{X}, j)$. Furthermore, let us define a function
\begin{equation}
    f_{maj}(x) = \begin{cases}
    1  & \text{if $x$ belongs to the majority class,} \\
    0  & \text{otherwise.}
\end{cases}
\end{equation}
We can then define the difficulty index of a collection of observations $\mathcal{X}$, parametrized by $m$, the number of considered nearest neighbors of each minority observation, as
\begin{equation}
    DI(\mathcal{X}, m) = \frac{1}{mn_{min}} \sum_{i=1}^{n_{min}}{\sum_{j=1}^{m}}{f_{maj}(f_{NN}(\mathcal{X}^{(i)}_{min}, \mathcal{X}, j))}.
    \label{eq:di}
\end{equation}

\begin{figure*}[!htb]
\centering
\begin{subfigure}[b]{0.22\textwidth}
  \includegraphics[width=\textwidth]{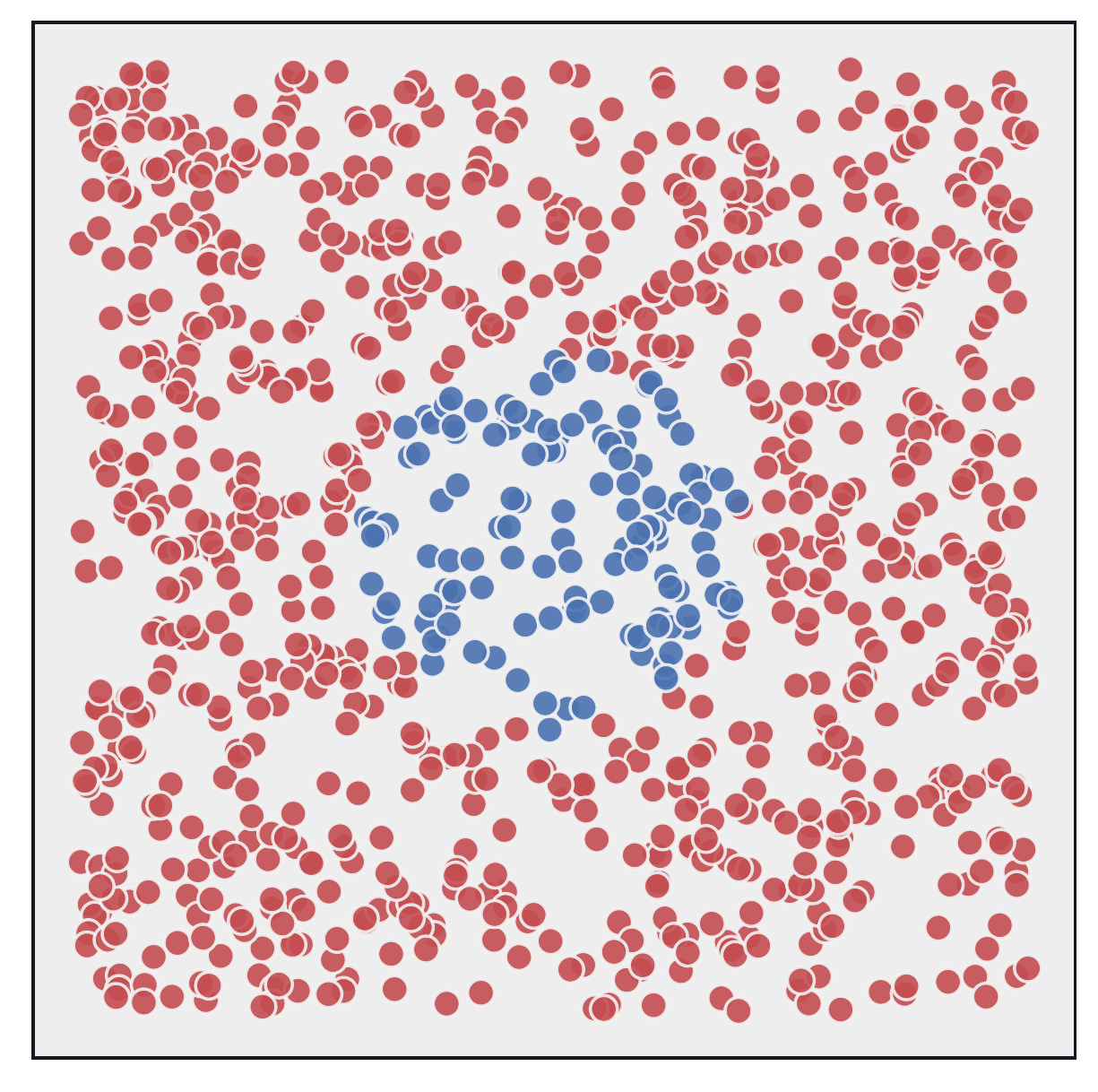}
  \caption{$DI = 0.058$}
\end{subfigure}
~
\begin{subfigure}[b]{0.22\textwidth}
  \includegraphics[width=\textwidth]{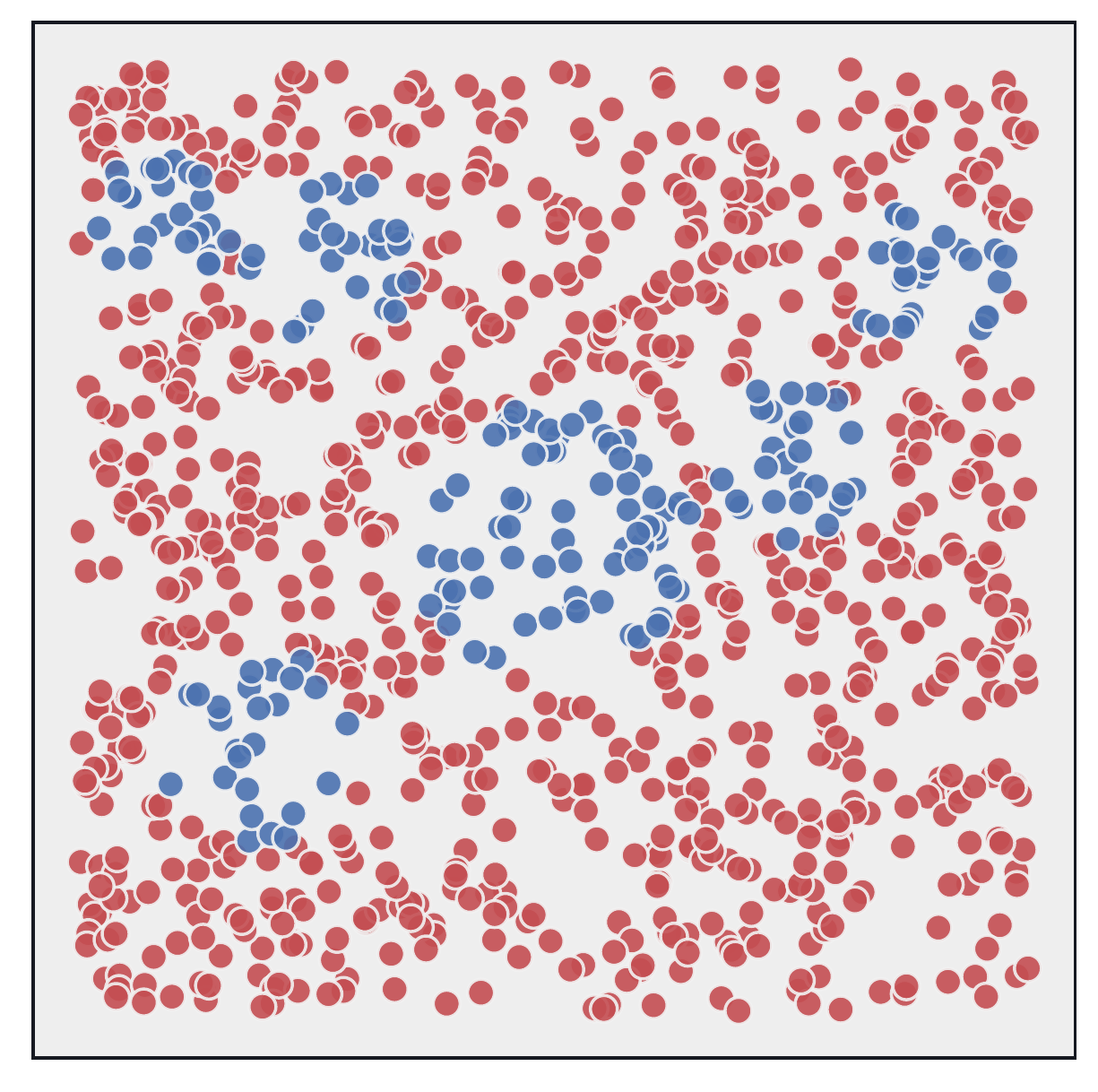}
  \caption{$DI = 0.185$}
\end{subfigure}
~
\begin{subfigure}[b]{0.22\textwidth}
  \includegraphics[width=\textwidth]{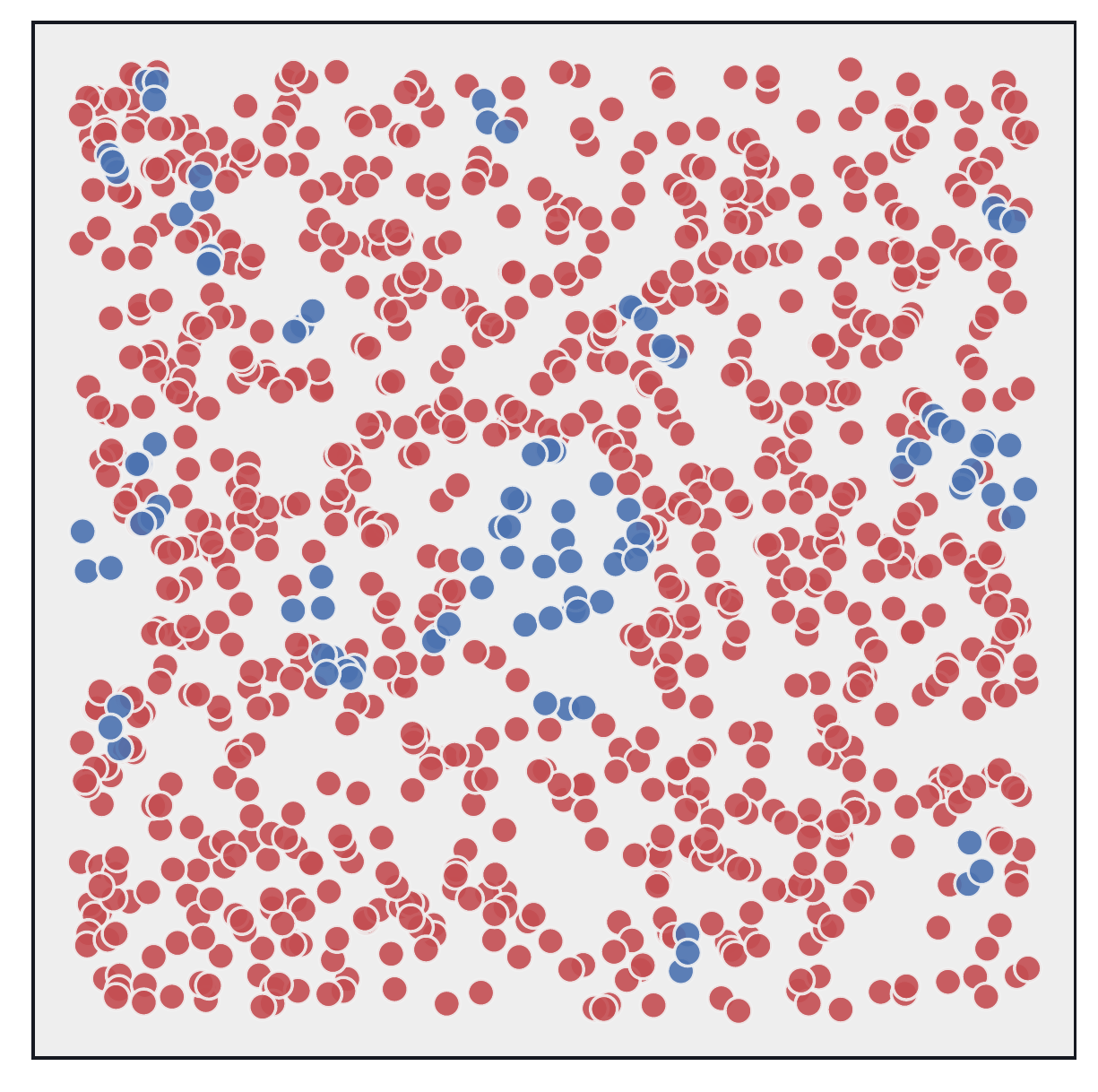}
  \caption{$DI = 0.461$}
\end{subfigure}
~
\begin{subfigure}[b]{0.22\textwidth}
  \includegraphics[width=\textwidth]{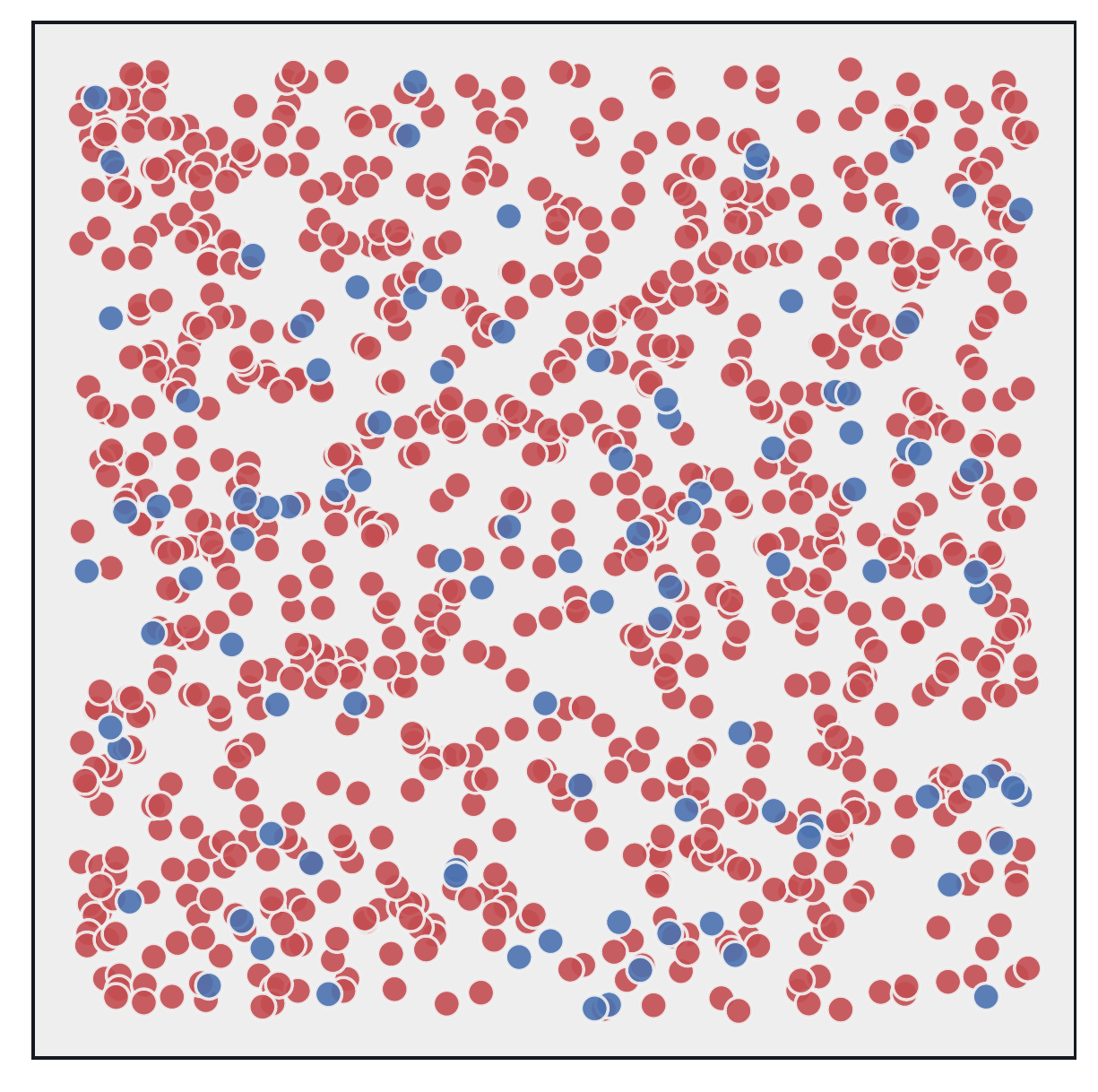}
  \caption{$DI = 0.886$}
\end{subfigure}
\caption{Example of increasingly more difficult imbalanced datasets and corresponding values of data difficulty index.}
\label{fig:example-di}
\end{figure*}

We present the impact of dataset characteristics on the calculated difficulty index in Figure~\ref{fig:example-di}. Similar to the original categorization, we used 5-neighborhood for its calculation. As can be seen, very low values of DI are obtained even for highly imbalanced datasets, as long as the class distributions can be clearly separated. However, as the entropy of the data increases, so do the obtained values of DI.

\section{Experimental study}
\label{sec:experimental-study}

To empirically evaluate the usefulness of the proposed PA algorithm we conducted a series of experiments, the aim of which was answering the following research questions:

\begin{itemize}
\item[RQ1:] How do PAO and PAU compare with the previously proposed radial-based resampling strategies?
\item[RQ2:] Is it possible to improve the individual performance of PAO and PAU by combining over- and undersampling?
\item[RQ3:] How does PA compare with state-of-the-art resampling strategies?
\item[RQ4:] Under what conditions does PA outperform other resampling algorithms?
\end{itemize}

\subsection{Set-up}

\noindent\textbf{Data.} Conducted experimental study was based on the binary imbalanced datasets provided in the KEEL repository \cite{alcala2011keel}, with a total of 60 datasets used. Their details were presented in Table~\ref{table:datasets}. In addition to the imbalance ratio (IR), the number of samples and the number of features, for each dataset we computed the data difficulty index (DI) using $m = 5$ nearest neighbors. Prior to resampling and classification, each dataset was preprocessed: categorical features were encoded as integers, and afterwards all features were standarized by removing the mean and scaling to unit variance.

\begin{sidewaystable*}
\caption{Summary of the characteristics of datasets used throughout the experimental study.}
\label{table:datasets}
\centering
\begin{tabular}{lrrrrlrrrr}
\toprule
Name & DI & IR & Samples & Features & Name & DI & IR & Samples & Features \\
\cmidrule(l){1-5} \cmidrule(l){6-10} 
kddcup-buffer\_overflow\_vs\_back & 0.10 & 73.43 & 2233 & 41 & cleveland-0\_vs\_4 & 0.60 & 12.31 & 173 & 13 \\
kddcup-rootkit-imap\_vs\_back & 0.17 & 100.14 & 2225 & 41 & poker-8-9\_vs\_6 & 0.64 & 58.40 & 1485 & 10 \\
ecoli2 & 0.20 & 5.46 & 336 & 7 & haberman & 0.67 & 2.78 & 306 & 3 \\
page-blocks0 & 0.22 & 8.79 & 5472 & 10 & yeast-0-5-6-7-9\_vs\_4 & 0.68 & 9.35 & 528 & 8 \\
page-blocks-1-3\_vs\_4 & 0.23 & 15.86 & 472 & 10 & zoo-3 & 0.68 & 19.20 & 101 & 16 \\
glass0 & 0.27 & 2.06 & 214 & 9 & yeast-0-3-5-9\_vs\_7-8 & 0.68 & 9.12 & 506 & 8 \\
ecoli-0-1\_vs\_2-3-5 & 0.28 & 9.17 & 244 & 7 & flare-F & 0.69 & 23.79 & 1066 & 11 \\
kr-vs-k-zero\_vs\_eight & 0.29 & 53.07 & 1460 & 6 & abalone-21\_vs\_8 & 0.70 & 40.50 & 581 & 8 \\
ecoli-0-1-4-7\_vs\_2-3-5-6 & 0.29 & 10.59 & 336 & 7 & yeast-1\_vs\_7 & 0.71 & 14.30 & 459 & 7 \\
ecoli1 & 0.33 & 3.36 & 336 & 7 & poker-8\_vs\_6 & 0.72 & 85.88 & 1477 & 10 \\
yeast3 & 0.35 & 8.10 & 1484 & 8 & poker-9\_vs\_7 & 0.75 & 29.50 & 244 & 10 \\
ecoli-0-6-7\_vs\_5 & 0.35 & 10.00 & 220 & 6 & yeast4 & 0.76 & 28.10 & 1484 & 8 \\
yeast-2\_vs\_4 & 0.39 & 9.08 & 514 & 8 & glass-0-1-4-6\_vs\_2 & 0.78 & 11.06 & 205 & 9 \\
glass1 & 0.40 & 1.82 & 214 & 9 & glass-0-1-6\_vs\_2 & 0.80 & 10.29 & 192 & 9 \\
ecoli-0-6-7\_vs\_3-5 & 0.40 & 9.09 & 222 & 7 & glass2 & 0.81 & 11.59 & 214 & 9 \\
yeast5 & 0.40 & 32.73 & 1484 & 8 & abalone9-18 & 0.82 & 16.40 & 731 & 8 \\
ecoli-0-2-6-7\_vs\_3-5 & 0.42 & 9.18 & 224 & 7 & yeast-1-2-8-9\_vs\_7 & 0.82 & 30.57 & 947 & 8 \\
glass-0-1-6\_vs\_5 & 0.42 & 19.44 & 184 & 9 & glass-0-1-5\_vs\_2 & 0.83 & 9.12 & 172 & 9 \\
ecoli-0-1-3-7\_vs\_2-6 & 0.43 & 39.14 & 281 & 7 & abalone-17\_vs\_7-8-9-10 & 0.84 & 39.31 & 2338 & 8 \\
yeast-2\_vs\_8 & 0.46 & 23.10 & 482 & 8 & yeast-1-4-5-8\_vs\_7 & 0.89 & 22.10 & 693 & 8 \\
vehicle1 & 0.48 & 2.90 & 846 & 18 & abalone-20\_vs\_8-9-10 & 0.89 & 72.69 & 1916 & 8 \\
pima & 0.48 & 1.87 & 768 & 8 & winequality-red-3\_vs\_5 & 0.90 & 68.10 & 691 & 11 \\
glass4 & 0.49 & 15.46 & 214 & 9 & winequality-red-4 & 0.92 & 29.17 & 1599 & 11 \\
ecoli3 & 0.50 & 8.60 & 336 & 7 & winequality-red-8\_vs\_6 & 0.92 & 35.44 & 656 & 11 \\
yeast-0-2-5-6\_vs\_3-7-8-9 & 0.51 & 9.14 & 1004 & 8 & winequality-white-3\_vs\_7 & 0.93 & 44.00 & 900 & 11 \\
glass5 & 0.51 & 22.78 & 214 & 9 & winequality-white-3-9\_vs\_5 & 0.94 & 58.28 & 1482 & 11 \\
yeast6 & 0.54 & 41.40 & 1484 & 8 & winequality-red-8\_vs\_6-7 & 0.94 & 46.50 & 855 & 11 \\
yeast1 & 0.54 & 2.46 & 1484 & 8 & abalone-19\_vs\_10-11-12-13 & 0.96 & 49.69 & 1622 & 8 \\
vehicle3 & 0.55 & 2.99 & 846 & 18 & poker-8-9\_vs\_5 & 0.97 & 82.00 & 2075 & 10 \\
winequality-white-9\_vs\_4 & 0.60 & 32.60 & 168 & 11 & abalone19 & 0.97 & 129.44 & 4174 & 8 \\
\bottomrule
\end{tabular}
\end{sidewaystable*}

\noindent\textbf{Classification.} Four different classification algorithms, representing different learning paradigms, were used throughout the experimental study: CART decision tree, k-nearest neighbors classifier (KNN), support vector machine (SVM) and multi-layer perceptron (MLP). The implementations of the classification algorithms provided in the scikit-learn machine learning library \cite{pedregosa2011scikit} were utilized. Used hyperparameters of the classification algorithms were presented in Table~\ref{table:parameters}.

\begin{table}
\begin{center}
\caption{Parameters of the classification and the sampling algorithms used throughout the experimental study.}
\label{table:parameters}
\begin{tabular}{ll}
\toprule
Algorithm & Parameters \\
\midrule
CART & criterion: Gini impurity \\
KNN & $k$-nearest neighbors = 3 \\
SVM & kernel: RBF; \\
 & C = 1.0 \\
MLP & hidden neurons = 100; \\
 & activation: ReLU; \\
 & optimizer: Adam; \\
 & learning rate = 0.001 \\
\midrule
PA & ratio = 0.1; \\
 & $\gamma$ = 0.5; \\
 & $\lambda$ = 10.0; \\
 & $n$ anchors = 10; \\
 & iterations = 200; \\
 & learning rate = 0.001 \\
SMOTE & $k$-nearest neighbors = 5 \\
pf-SMOTE & topology: star \\
Lee & $k$-nearest neighbors = 5; \\
 & rejection level = 0.5 \\
SMOBD & $\eta_1$ = 0.5; \\
 & noise threshold $t$ = 1.8 \\
G-SMOTE & $k$-nearest neighbors = 5 \\
LVQ-SMOTE & $k$-nearest neighbors = 5; \\
 & $n$ clusters = 10 \\
A-SMOTE & $k$-nearest neighbors = 5; \\
 & population parameter = 2 \\
SMOTE-TL & $k$-nearest neighbors = 5 \\
RBO & $\gamma$ = 0.05; \\
 & step size = 0.001; \\
 & $n$ steps = 500 \\
RBU & $\gamma$ = 0.05 \\
\bottomrule
\end{tabular}
\end{center}
\end{table}

\noindent\textbf{Reference resampling methods.} In addition to two previously proposed methods utilizing the concept of class potential, Radial-Based Oversampling (RBO) \cite{koziarski2019radial} and Radial-Based Undersampling (RBU) \cite{koziarski2020radial}, we considered several other state-of-the-art resampling strategies. We based our choice on a recent ranking constructed by Kovács \cite{kovacs2019empirical}, out of which we selected the following best-performing methods: SMOTE \cite{chawla2002smote}, Polynomial Fitting SMOTE (pf-SMOTE) \cite{gazzah2008new}, Oversampling with Rejection (Lee) \cite{lee2015over}, Synthetic Minority Oversampling Based on Sample Density (SMOBD) \cite{cao2011applying}, Partially Guided Oversampling (G-SMOTE) \cite{sandhan2014handling}, Learning Vector Quantization-based SMOTE (LVQ-SMOTE) \cite{nakamura2013lvq}, Assembled SMOTE (A-SMOTE) \cite{zhou2013quasi} and SMOTE combined with Tomek Links (SMOTE-TL) \cite{batista2004study}. With the exception of RBO and RBU, the implementations of the reference methods provided in the smote-variants library \cite{kovacs2019smote} were utilized. Used hyperparameters of the resampling algorithms were presented in Table~\ref{table:parameters}. Throughout the experimental study all resamplers were used up to the point of achieving a balanced class distribution.

\noindent\textbf{Evaluation.} For every dataset we reported the results averaged over the $5\times2$ cross-validation folds \cite{alpaydin1999combined}. Throughout the experimental study we reported the values of precision, recall, AUC and G-mean. We intentionally excluded F-measure from the considered performance metrics, since it was previously shown \cite{brzezinski2019dynamics} that F-measure is usually more biased towards the majority class than AUC and G-mean, making AUC and G-mean more suitable for assessment of performance in the imbalanced data classification task

\noindent\textbf{Implementation and reproducibility.} The experiments described in this paper were implemented in the Python programming language. Complete code, sufficient to repeat the experiments, was made publicly available at\footnote{\url{https://github.com/michalkoziarski/PotentialAnchoring}}. In addition to the code we also provided the cross-validation folds used during the experiments, as well as the raw result files, which can be used for further analysis.

\subsection{Comparison with previous radial-based strategies}

\begin{figure*}[!htb]
\centering
\begin{subfigure}[b]{0.22\textwidth}
  \includegraphics[width=\textwidth]{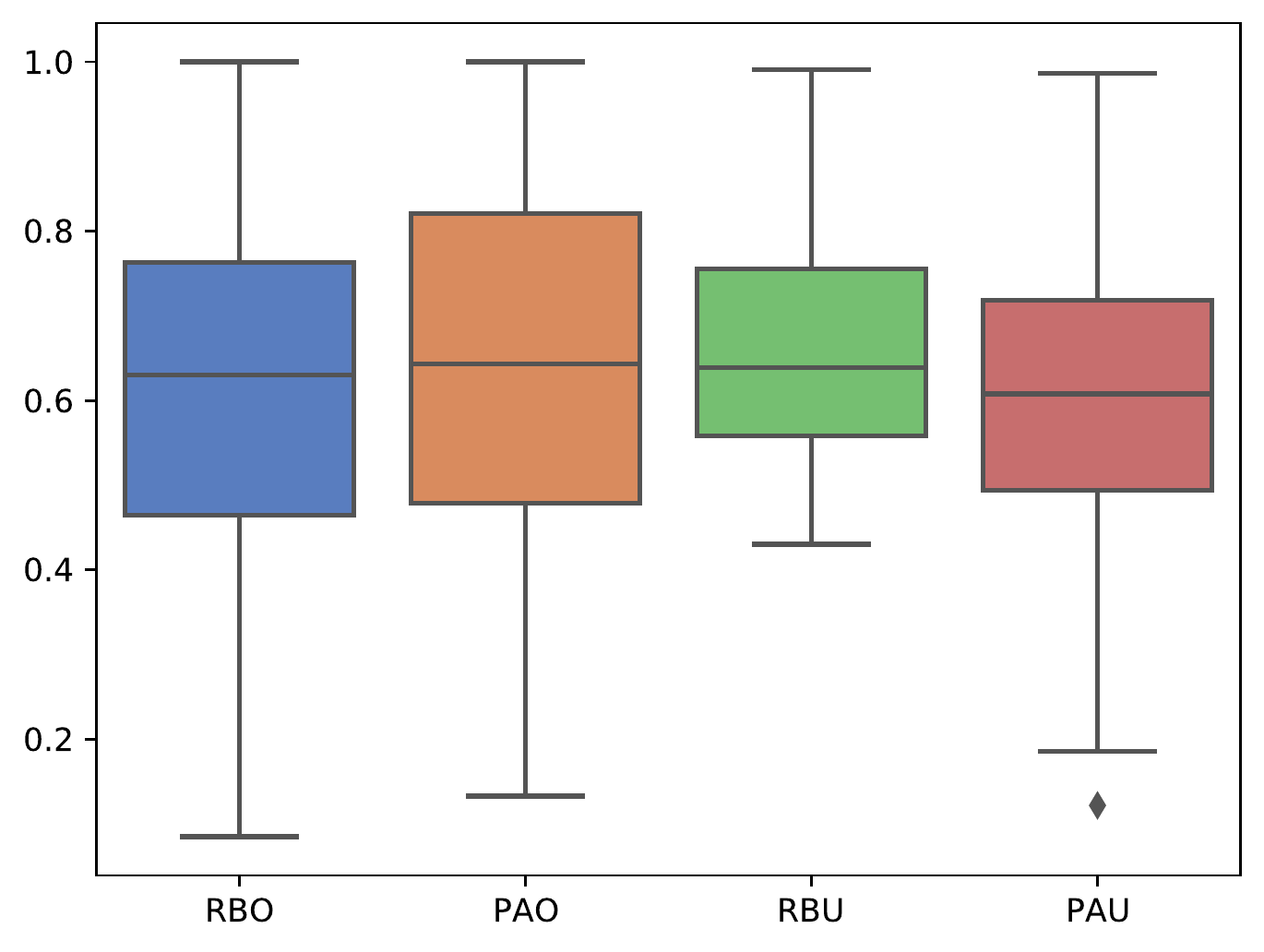}
  \caption{CART}
\end{subfigure}
~
\begin{subfigure}[b]{0.22\textwidth}
  \includegraphics[width=\textwidth]{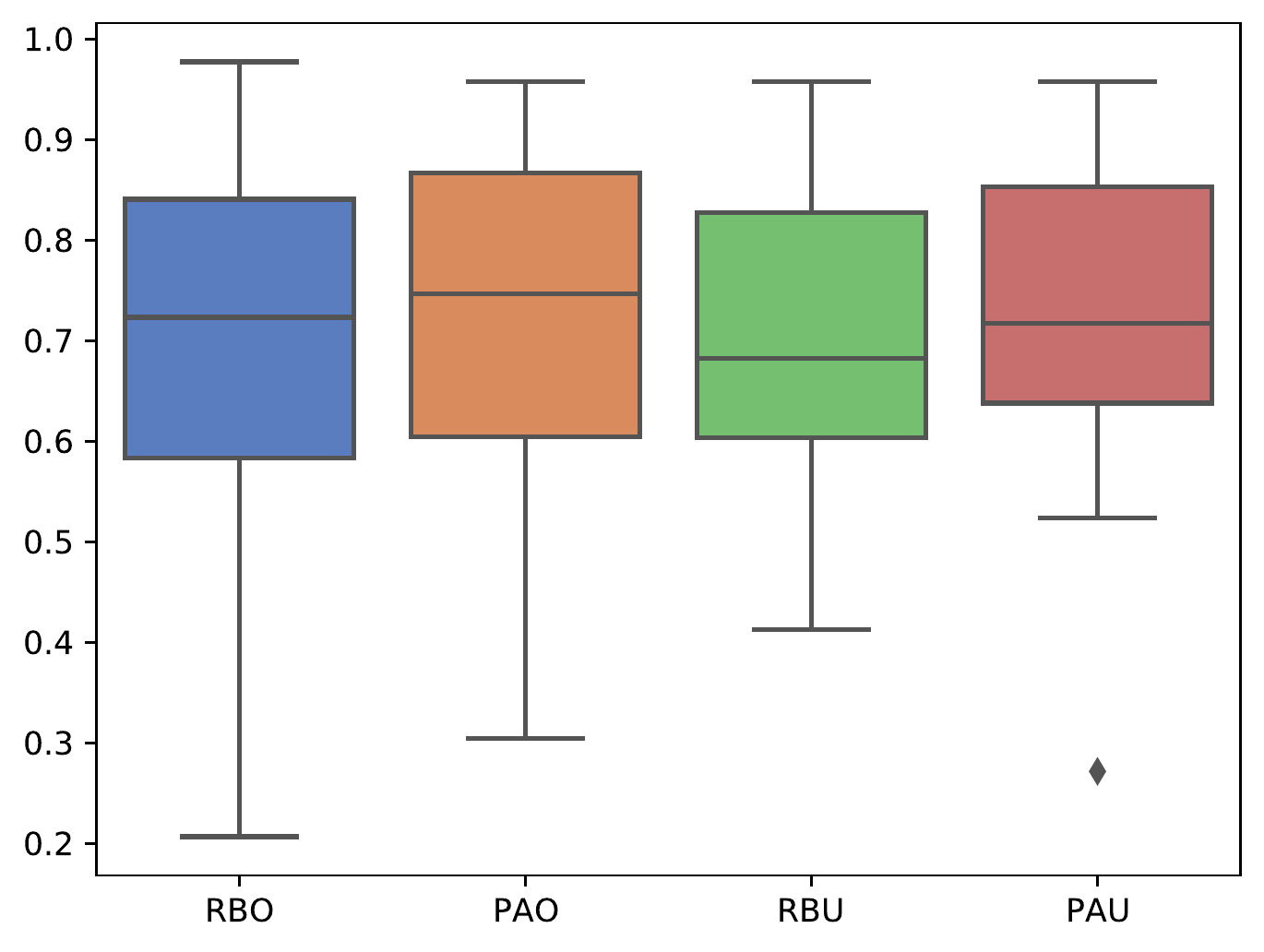}
  \caption{KNN}
\end{subfigure}
~
\begin{subfigure}[b]{0.22\textwidth}
  \includegraphics[width=\textwidth]{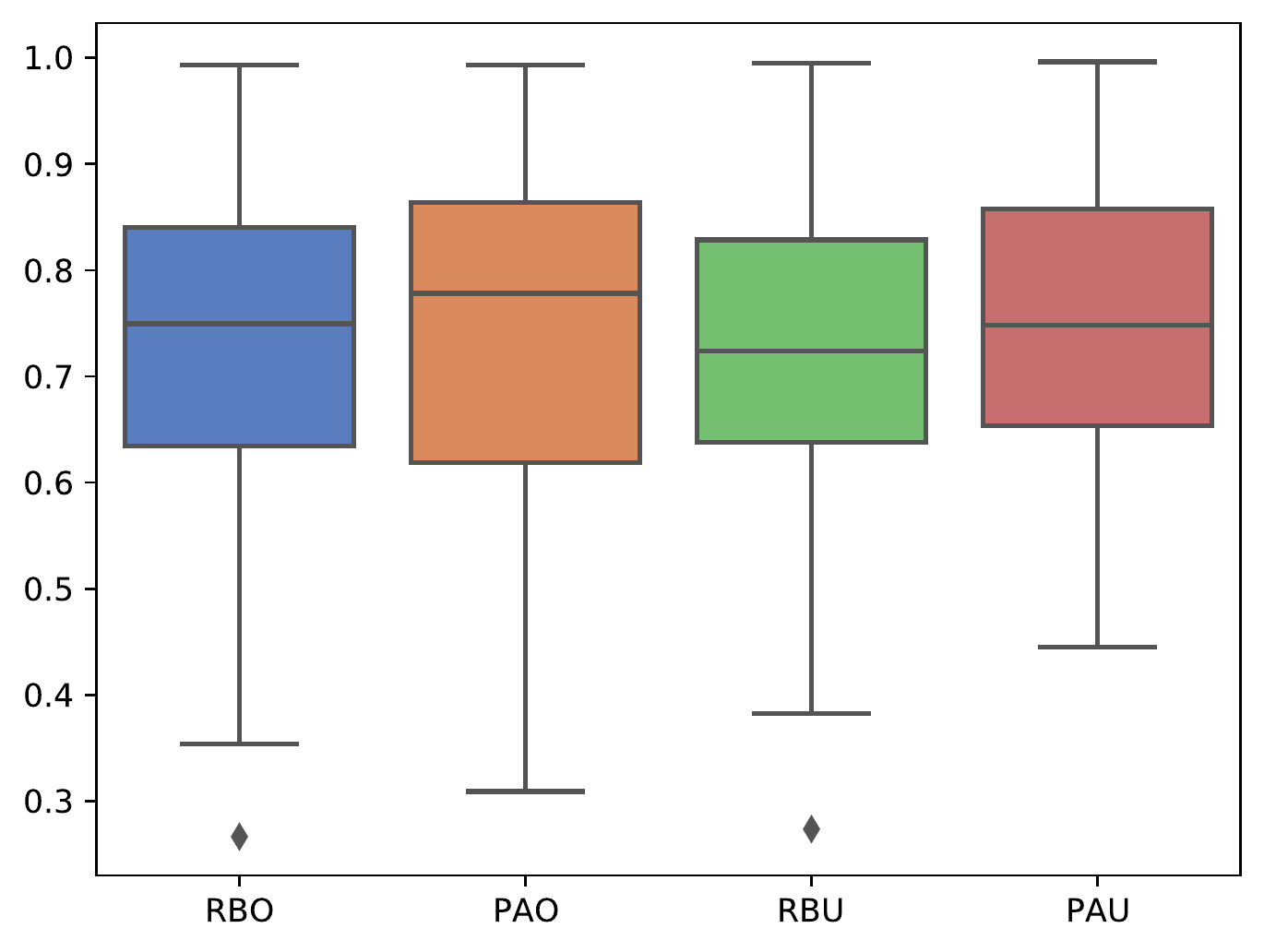}
  \caption{SVM}
\end{subfigure}
~
\begin{subfigure}[b]{0.22\textwidth}
  \includegraphics[width=\textwidth]{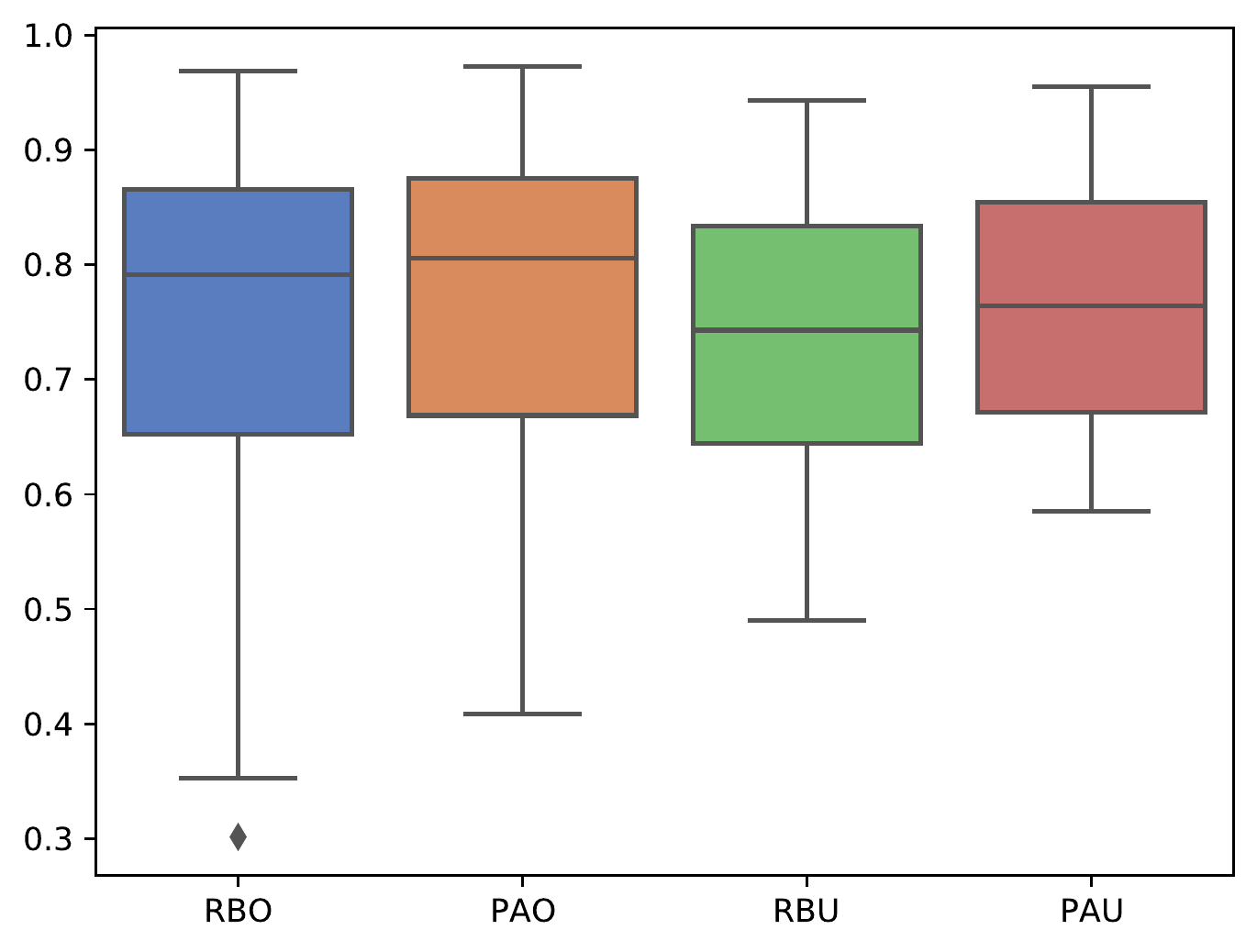}
  \caption{MLP}
\end{subfigure}
\caption{Comparison of classification performance of RBO, PAO, RBU and PAU with respect to G-mean, averaged over all of the considered datasets.}
\label{fig:pa-rb-boxplots}
\end{figure*}

As previously discussed, similar to RBO and RBU algorithms, PA relies on the concept of class potential to estimate the local density of observations from a given class. Even though both families of algorithms use class potential differently during the resampling process, a natural question is whether the approach proposed in this paper, by itself, improves the performance during both over- and undersampling. To answer this question we began our analysis with a pairwise comparison of both PAO and RBO, as well as PAU and RBU. To asses the statistical significance of this comparison we conducted a two-sided Wilcoxon signed-rank test, the results of which were presented in Table~\ref{table:pa-rb-comparison-oversampling} for the pair of oversampling algorithms, and in Table~\ref{table:pa-rb-comparison-undersampling} for the pair of undersampling algorithms. Furthermore, boxplots with respect to G-mean achieved by all four methods were presented in Figure~\ref{fig:pa-rb-boxplots}.

\begin{table}
\small
\caption{Results of the comparison between PAO and RBO oversampling. The number of datasets on which given method achieved better performance and $p$-value obtained during comparison with Wilcoxon signed-rank test.}
\label{table:pa-rb-comparison-oversampling}
\centering
\begin{tabular}{lllllll}
\toprule
& \multicolumn{3}{l}{AUC} & \multicolumn{3}{l}{G-mean} \\
\cmidrule(l){2-4} \cmidrule(l){5-7}
& RBO & PAO & $p$-value & RBO & PAO & $p$-value \\
\midrule
CART & 17 & 39 & \textbf{0.0002} & 16 & 42 & \textbf{0.0000} \\
KNN & 16 & 42 & \textbf{0.0004} & 18 & 41 & \textbf{0.0000} \\
SVM & 11 & 48 & \textbf{0.0000} & 10 & 49 & \textbf{0.0000} \\
MLP & 14 & 44 & \textbf{0.0000} & 11 & 47 & \textbf{0.0000} \\
\bottomrule
\end{tabular}
\end{table}

\begin{table}
\small
\caption{Results of the comparison between PAU and RBU undersampling. The number of datasets on which given method achieved better performance and $p$-value obtained during comparison with Wilcoxon signed-rank test.}
\label{table:pa-rb-comparison-undersampling}
\centering
\begin{tabular}{lllllll}
\toprule
& \multicolumn{3}{l}{AUC} & \multicolumn{3}{l}{G-mean} \\
\cmidrule(l){2-4} \cmidrule(l){5-7}
& RBU & PAU & $p$-value & RBU & PAU & $p$-value \\
\midrule
CART & 32 & 28 & 0.1976 & 38 & 22 & \textbf{0.0007} \\
KNN & 23 & 37 & \textbf{0.0088} & 22 & 37 & \textbf{0.0063} \\
SVM & 23 & 37 & \textbf{0.0030} & 21 & 39 & \textbf{0.0013} \\
MLP & 16 & 44 & \textbf{0.0004} & 17 & 43 & \textbf{0.0001} \\
\bottomrule
\end{tabular}
\end{table}

As can be seen, with the sole exception of AUC of undersampling algorithms combined with the CART classifier, the results of this pairwise comparison were statistically significant. Furthermore, with the only exception of the comparison of undersampling algorithms combined with the CART classifier (for both AUC and G-mean), resampling strategies based on the potential anchoring paradigm outperformed the radial-based strategies, as demonstrated by a better average performance and a higher number of datasets on which potential anchoring achieved better results. It is also worth noting that the improvement in performance was stronger in the case of oversampling than in the case of undersampling, as evidenced by lower $p$-values and higher number of datasets on which PAO outperformed RBO. In general, the observed results lead to a conclusion that potential anchoring paradigm, on average, produces an improvement in performance compared to the radial-based strategies, both for over- and undersampling.

\subsection{Combining over- and undersampling}

The second extension of previous radial-based approaches proposed in this paper is combining over- and undersampling with the goal of improving the performance of individual algorithms. To evaluate the hypothesis that using the combined over- and undersampling leads to an improved performance we conducted an experiment in which we modulated the ratio of imbalance alleviated by oversampling. Specifically, we considered the values of ratio parameter $\in \{0.0, 0.1, 0.2, ..., 1.0\}$, with higher values of ratio parameter corresponding to stronger oversampling and weaker undersampling, both applied together up to the point of achieving balanced class distribution. In particular, ratio equal to 1 indicated that only the oversampling was used (or, in other words, PA degenerated to PAO), and ratio equal to 0 indicated that only the undersampling was used (PA degenerated to PAU).

\begin{figure*}
\centering
\includegraphics[width=0.7\textwidth]{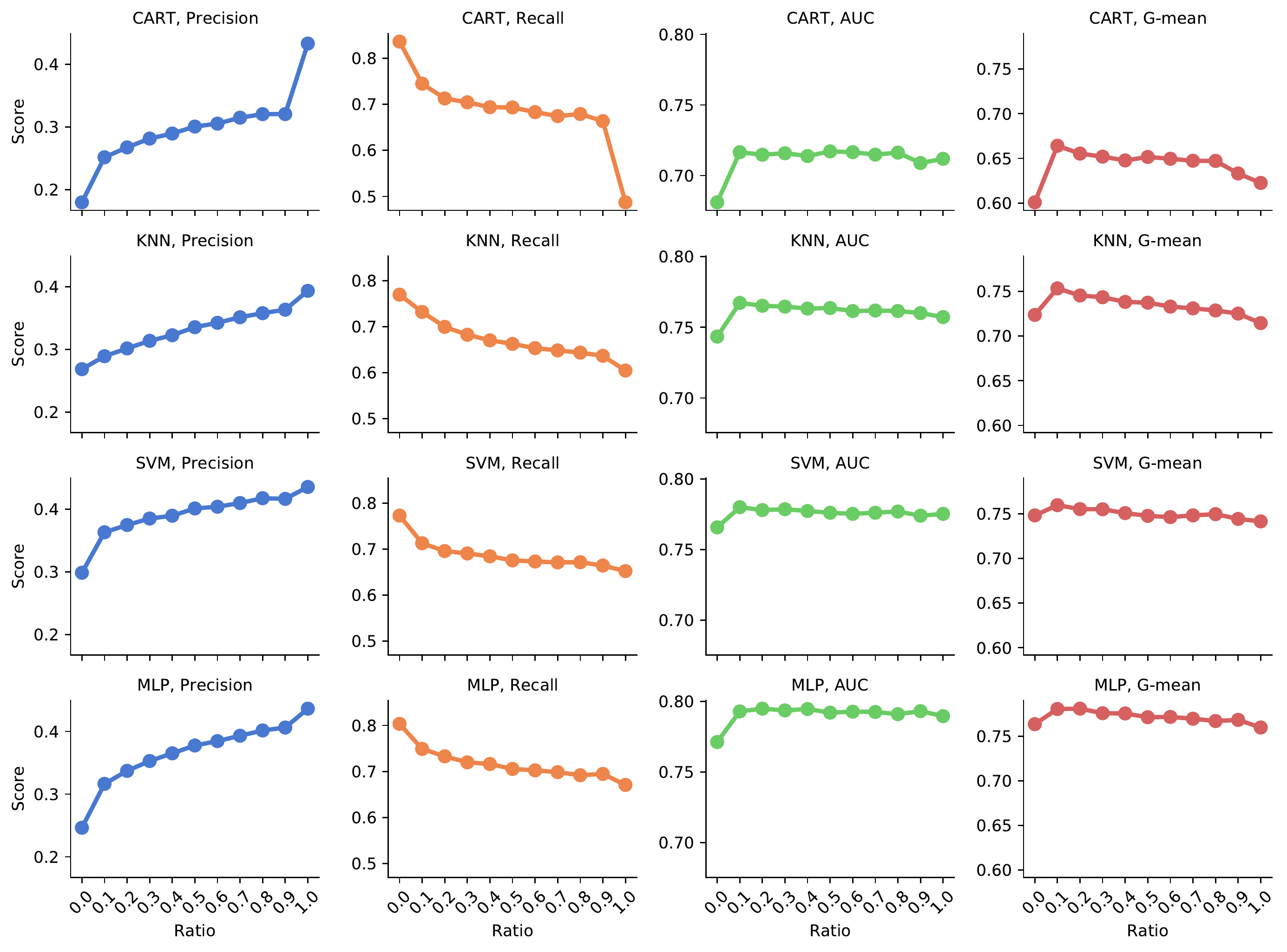}
\caption{Visualization of the impact of ratio parameter on average classification performance. Ratio equal to 0 indicates the case in which class balancing is done solely via undersampling (PAU), whereas ratio equal to 1 indicates the case in which it is done solely via oversampling (PAO).}
\label{fig:ratio}
\end{figure*}

The results, averaged over all of the datasets, were presented in Figure~\ref{fig:ratio}. First of all, as can be seen, the ratio impacted both precision and recall monotonically, with the precision increasing with the increase of the ratio, and the recall decreasing. In other words, stronger oversampling produced better precision of the predictions at the cost of their recall, and stronger undersampling had the opposite effect. This trend was consistent for all of the considered classification algorithms. Secondly, perhaps more importantly, the ratio also affected the performance measured with respect to the combined metrics, that is AUC and G-mean. However, contrary to the results observed for precision and recall, this trend was not monotonic across all of the considered ratio values, but instead displayed a peak at the ratio equal to 0.1, shared for all of the considered classifiers and both combined metrics. This was the case regardless of the exact shape of the performance curve in any specific case, and in particular regardless of which of the PAO and PAU displayed a better average performance for a considered classifier and metric combination. This leads to a conclusion that the strategy of eliminating the imbalance by a combination of over- and undersampling is beneficial to the performance of techniques based on the potential anchoring paradigm when strong undersampling is combined with a weak oversampling. This is likely a case due to achieving better precision-recall trade-off with respect to the combined metrics.

\subsection{Comparison with reference resampling strategies}
\label{sec:reference}

\begin{figure*}[!htb]
\centering
\begin{subfigure}[b]{0.4\textwidth}
  \includegraphics[width=\textwidth]{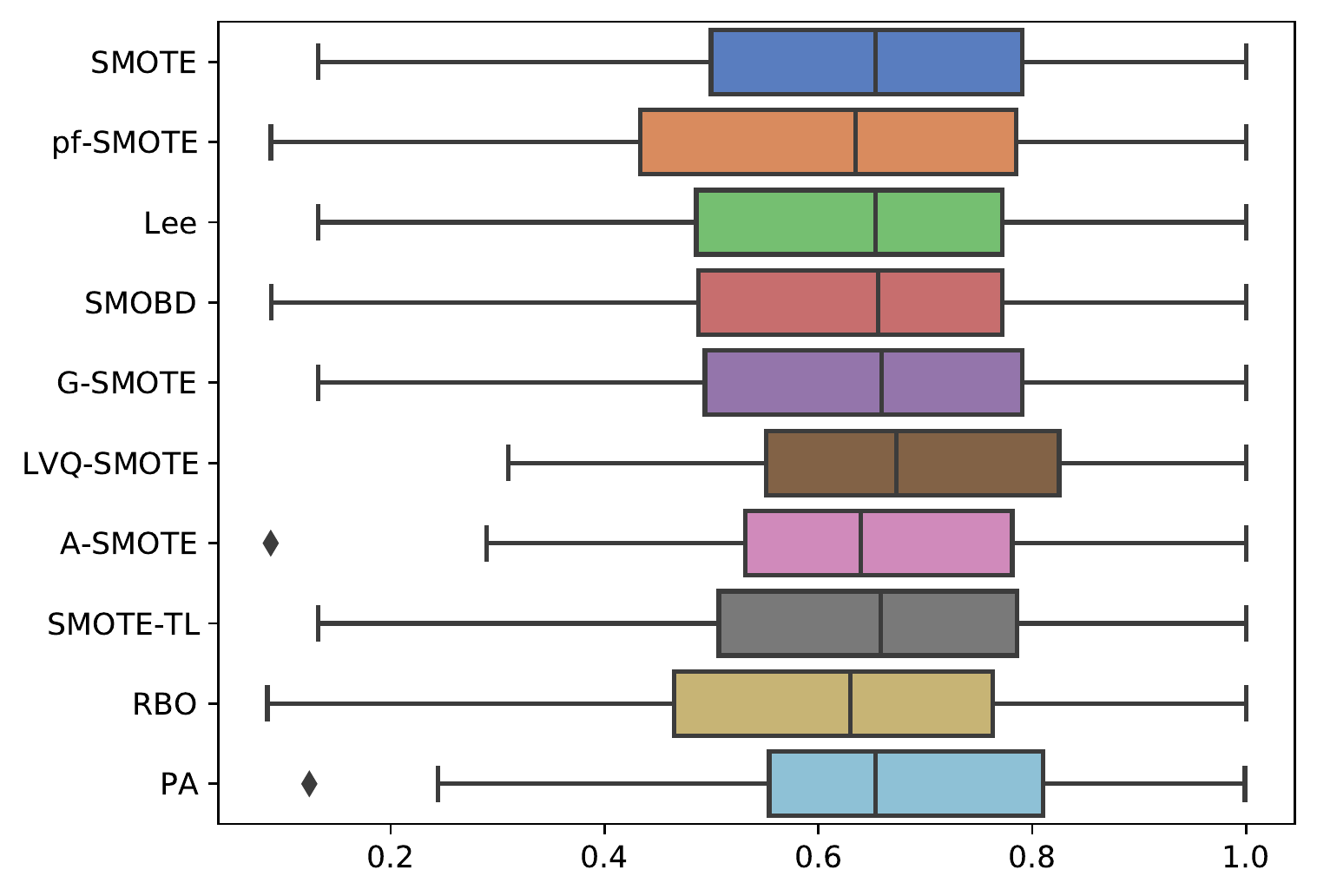}
  \caption{CART}
\end{subfigure}
~
\begin{subfigure}[b]{0.4\textwidth}
  \includegraphics[width=\textwidth]{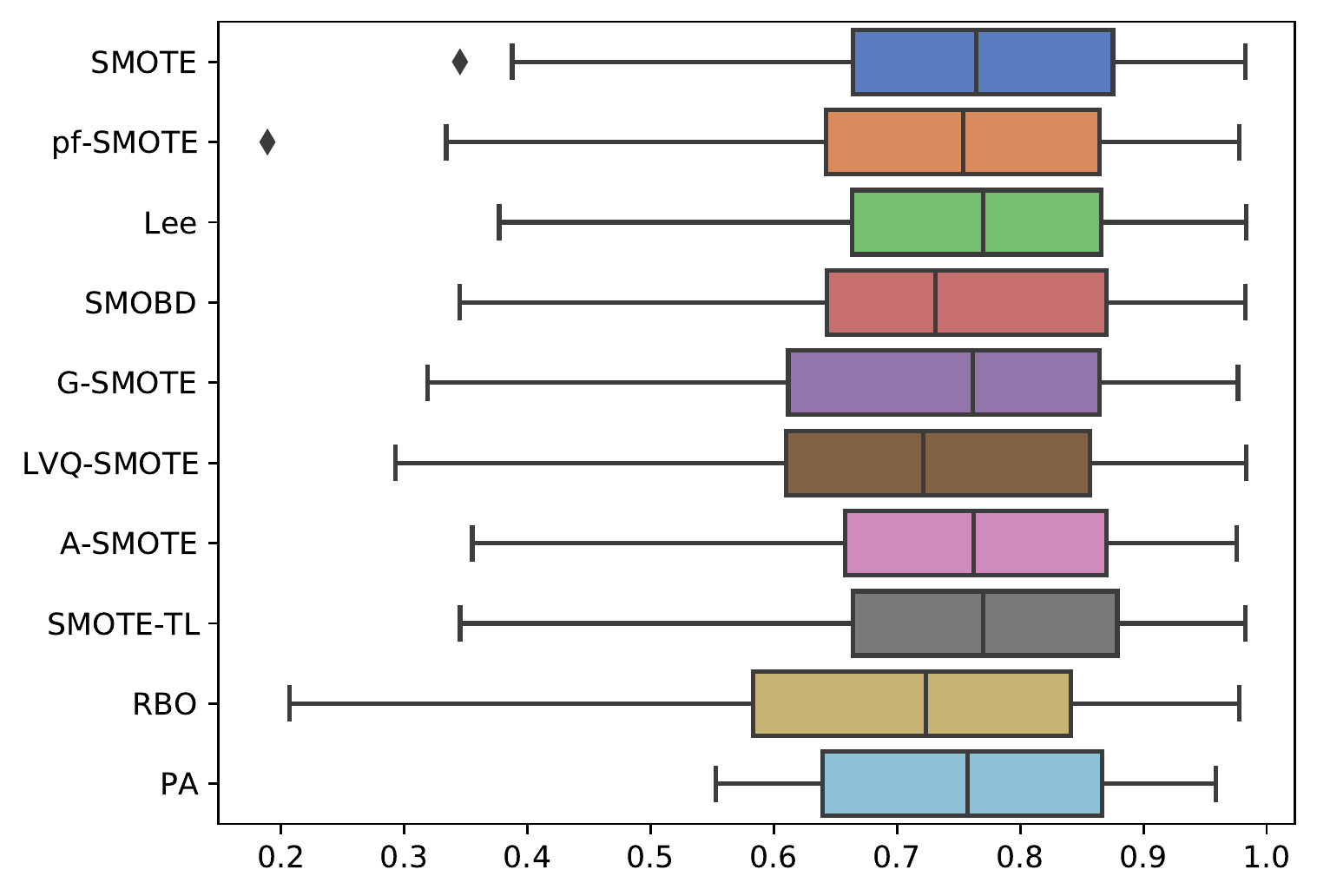}
  \caption{KNN}
\end{subfigure}

\begin{subfigure}[b]{0.4\textwidth}
  \includegraphics[width=\textwidth]{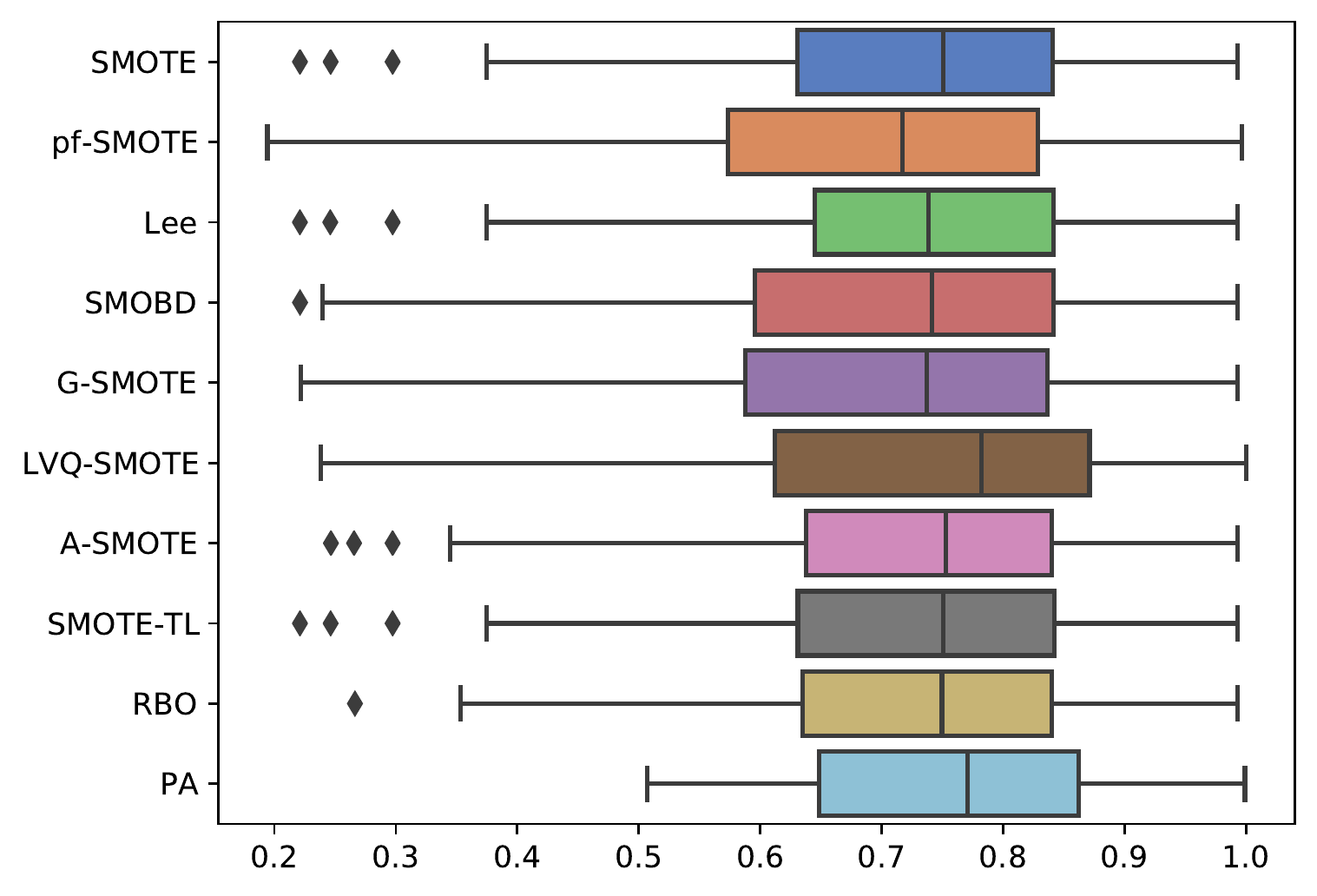}
  \caption{SVM}
\end{subfigure}
~
\begin{subfigure}[b]{0.4\textwidth}
  \includegraphics[width=\textwidth]{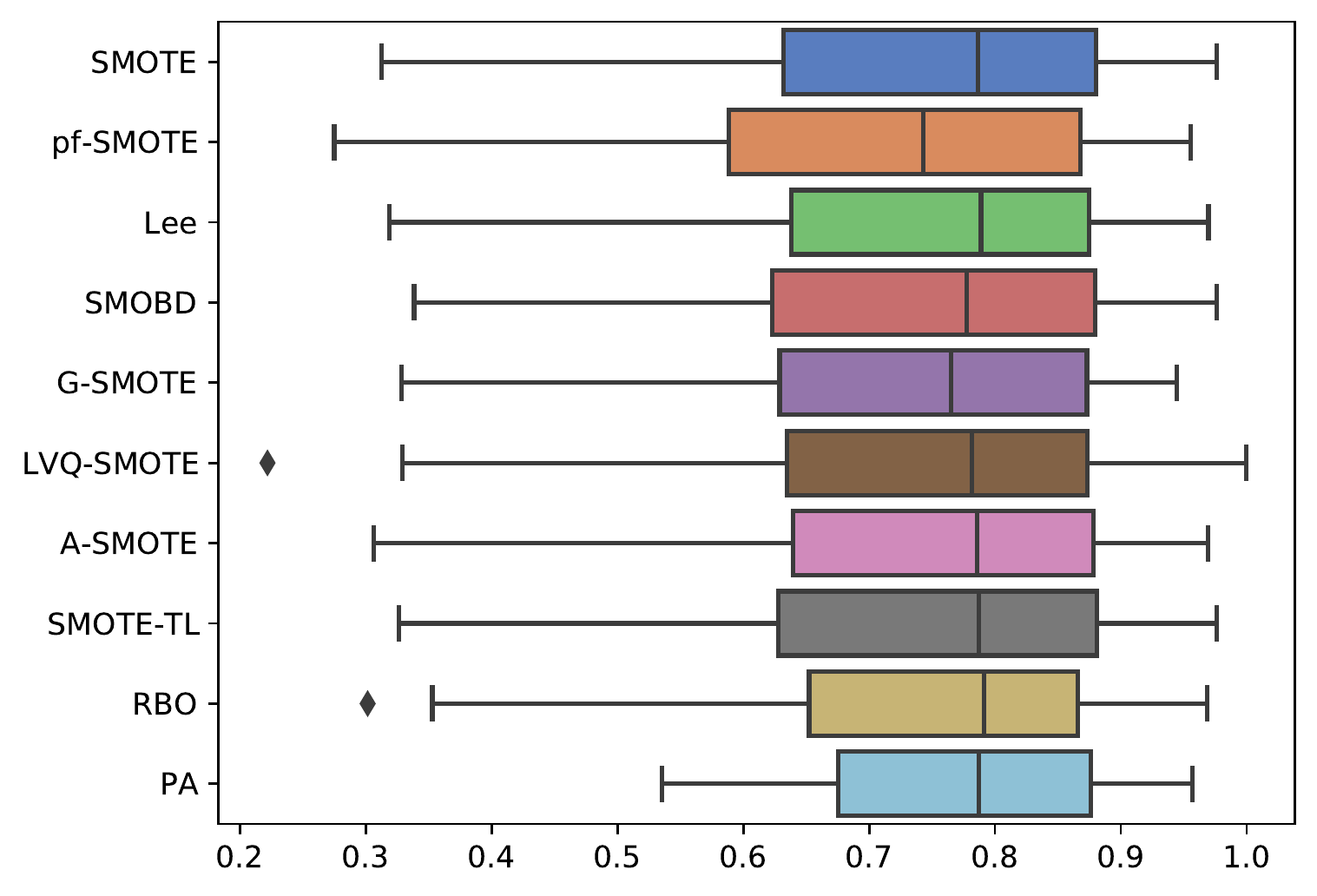}
  \caption{MLP}
\end{subfigure}
\caption{Comparison of average performance of different resampling strategies with respect to G-mean, calculated for all of the considered datasets.}
\label{fig:final-boxplots-all}
\end{figure*}

In the next stage of the experimental study we compared the performance of PA to that of the reference methods. To assess the statistical significance of this comparison we employed Friedman test combined with Shaffer's post-hoc. We reported the results at the significance level $\alpha = 0.10$. We present the summary of this statistical comparison, containing average ranks achieved by each method as well as the indication of the cases in which statistically significant differences were observed, in Table~\ref{table:results-final}. Furthermore, we also present boxplots illustrating the performance of each method with respect to G-mean in Figure~\ref{fig:final-boxplots-all}.

\begin{table*}
\scriptsize
\caption{Average ranks of the evaluated methods calculated for all of the considered datasets. Best performance was denoted with bold font. Methods that achieved significantly different results (according to Shaffer's post-hoc test) than PA where denoted in subscript: with $+$ sign for methods compared to which PA achieved a better results, and $-$ sign for methods compared to which PA achieved worse results.}
\label{table:results-final}
\centering
\begin{tabularx}{\textwidth}{llllllllllll}
\toprule
 & Metric & SMOTE & pf-SMOTE & Lee & SMOBD & G-SMOTE & LVQ-SMOTE & A-SMOTE & SMOTE-TL & RBO & PA \\
\midrule
\multirow{4}{*}{CART} & Precision & 5.14 \textsubscript{--} & 4.40 \textsubscript{--} & 5.73 \textsubscript{--} & 5.22 \textsubscript{--} & 4.40 \textsubscript{--} & 7.17 \textsubscript{--} & 5.42 \textsubscript{--} & 4.78 \textsubscript{--} & \textbf{2.98} \textsubscript{--} & 9.77 \\
 & Recall & 5.05 \textsubscript{+} & 7.82 \textsubscript{+} & 6.01 \textsubscript{+} & 5.88 \textsubscript{+} & 6.48 \textsubscript{+} & 3.75 \textsubscript{+} & 5.47 \textsubscript{+} & 5.08 \textsubscript{+} & 7.92 \textsubscript{+} & \textbf{1.52} \\
 & G-mean & 4.83 & 7.48 \textsubscript{+} & 5.98 \textsubscript{+} & 5.74 \textsubscript{+} & 6.10 \textsubscript{+} & \textbf{3.80} & 5.18 & 4.68 & 7.18 \textsubscript{+} & 4.03 \\
 & AUC & 4.99 & 7.04 \textsubscript{+} & 5.97 & 5.77 & 5.65 & \textbf{4.08} & 5.32 & 4.61 & 6.88 \textsubscript{+} & 4.68 \\
\midrule
\multirow{4}{*}{KNN} & Precision & 5.69 \textsubscript{--} & \textbf{2.98} \textsubscript{--} & 5.70 \textsubscript{--} & 5.87 \textsubscript{--} & 3.67 \textsubscript{--} & 6.11 \textsubscript{--} & 5.58 \textsubscript{--} & 5.80 \textsubscript{--} & 3.92 \textsubscript{--} & 9.68 \\
 & Recall & 4.11 \textsubscript{+} & 6.81 \textsubscript{+} & 4.69 \textsubscript{+} & 4.72 \textsubscript{+} & 7.56 \textsubscript{+} & 7.36 \textsubscript{+} & 4.88 \textsubscript{+} & 4.03 \textsubscript{+} & 8.86 \textsubscript{+} & \textbf{1.99} \\
 & G-mean & 4.19 & 5.29 & 4.55 & 4.68 & 6.78 \textsubscript{+} & 7.03 \textsubscript{+} & 4.99 & \textbf{3.96} & 8.48 \textsubscript{+} & 5.05 \\
 & AUC & 4.20 & 5.04 & 4.55 & 4.68 & 6.37 & 7.06 & 5.08 & \textbf{4.06} \textsubscript{--} & 8.28 \textsubscript{+} & 5.69 \\
\midrule
\multirow{4}{*}{SVM} & Precision & 5.38 \textsubscript{--} & \textbf{2.42} \textsubscript{--} & 5.39 \textsubscript{--} & 5.03 \textsubscript{--} & 4.53 \textsubscript{--} & 5.92 \textsubscript{--} & 5.79 \textsubscript{--} & 5.49 \textsubscript{--} & 6.18 \textsubscript{--} & 8.87 \\
 & Recall & 5.68 \textsubscript{+} & 8.93 \textsubscript{+} & 5.50 \textsubscript{+} & 5.58 \textsubscript{+} & 7.14 \textsubscript{+} & 4.03 \textsubscript{+} & 5.67 \textsubscript{+} & 5.55 \textsubscript{+} & 4.88 \textsubscript{+} & \textbf{2.04} \\
 & G-mean & 5.33 \textsubscript{+} & 8.16 \textsubscript{+} & 5.48 \textsubscript{+} & 5.39 \textsubscript{+} & 6.82 \textsubscript{+} & 3.90 & 5.93 \textsubscript{+} & 5.48 \textsubscript{+} & 5.09 \textsubscript{+} & \textbf{3.42} \\
 & AUC & 5.43 \textsubscript{+} & 7.89 \textsubscript{+} & 5.67 \textsubscript{+} & 5.42 \textsubscript{+} & 6.52 \textsubscript{+} & 3.80 & 5.92 \textsubscript{+} & 5.54 \textsubscript{+} & 5.14 & \textbf{3.67} \\
\midrule
\multirow{4}{*}{MLP} & Precision & 5.51 \textsubscript{--} & \textbf{3.08} \textsubscript{--} & 5.20 \textsubscript{--} & 4.62 \textsubscript{--} & 4.15 \textsubscript{--} & 6.67 \textsubscript{--} & 5.23 \textsubscript{--} & 5.16 \textsubscript{--} & 5.40 \textsubscript{--} & 9.98 \\
 & Recall & 5.54 \textsubscript{+} & 8.93 \textsubscript{+} & 5.23 \textsubscript{+} & 5.98 \textsubscript{+} & 7.08 \textsubscript{+} & 4.65 \textsubscript{+} & 5.52 \textsubscript{+} & 5.56 \textsubscript{+} & 4.91 \textsubscript{+} & \textbf{1.60} \\
 & G-mean & 5.36 \textsubscript{+} & 8.55 \textsubscript{+} & 5.05 \textsubscript{+} & 5.38 \textsubscript{+} & 6.60 \textsubscript{+} & 5.00 & 5.42 \textsubscript{+} & 5.54 \textsubscript{+} & 4.68 & \textbf{3.42} \\
 & AUC & 5.56 \textsubscript{+} & 8.08 \textsubscript{+} & 5.15 & 5.41 & 6.38 \textsubscript{+} & 4.93 & 5.50 \textsubscript{+} & 5.59 \textsubscript{+} & 4.60 & \textbf{3.80} \\
\bottomrule
\end{tabularx}
\end{table*}

Several observations can be made based on the presented results. First of all, compared to the reference methods PA achieved significantly better recall at the cost of significantly worse precision, indicating stronger bias towards the minority class than the reference methods. When combined performance metrics were considered, PA achieved particularly strong performance when combined with SVM and MLP classifiers, in both cases achieving the highest average ranks for both AUC and G-mean. This outperformance was statistically significant: in the case of SVM, PA achieved significantly better results with respect to at least one of the combined metrics in 8 out of 9 cases (with the exception of LVQ-SMOTE), and in the case of MLP in 7 out of 9 cases (with the exception of LVQ-SMOTE and RBO). Statistically significant outperformance was also observed, to a lesser extent, in the case of CART and KNN classifiers, when compared to 5 reference methods for CART and 3 reference methods for KNN. Importantly, only in a single case of AUC measured for CART classifier did PA achieve a significantly worse performance than the reference method. Finally, as can be seen on the presented boxplots, PA tended to achieve visibly higher minimum performance than the reference methods, indicating its usefulness in a general case.

\subsection{Examination of factors influencing the performance of PA}

\begin{figure*}
\centering
\includegraphics[width=0.6\textwidth]{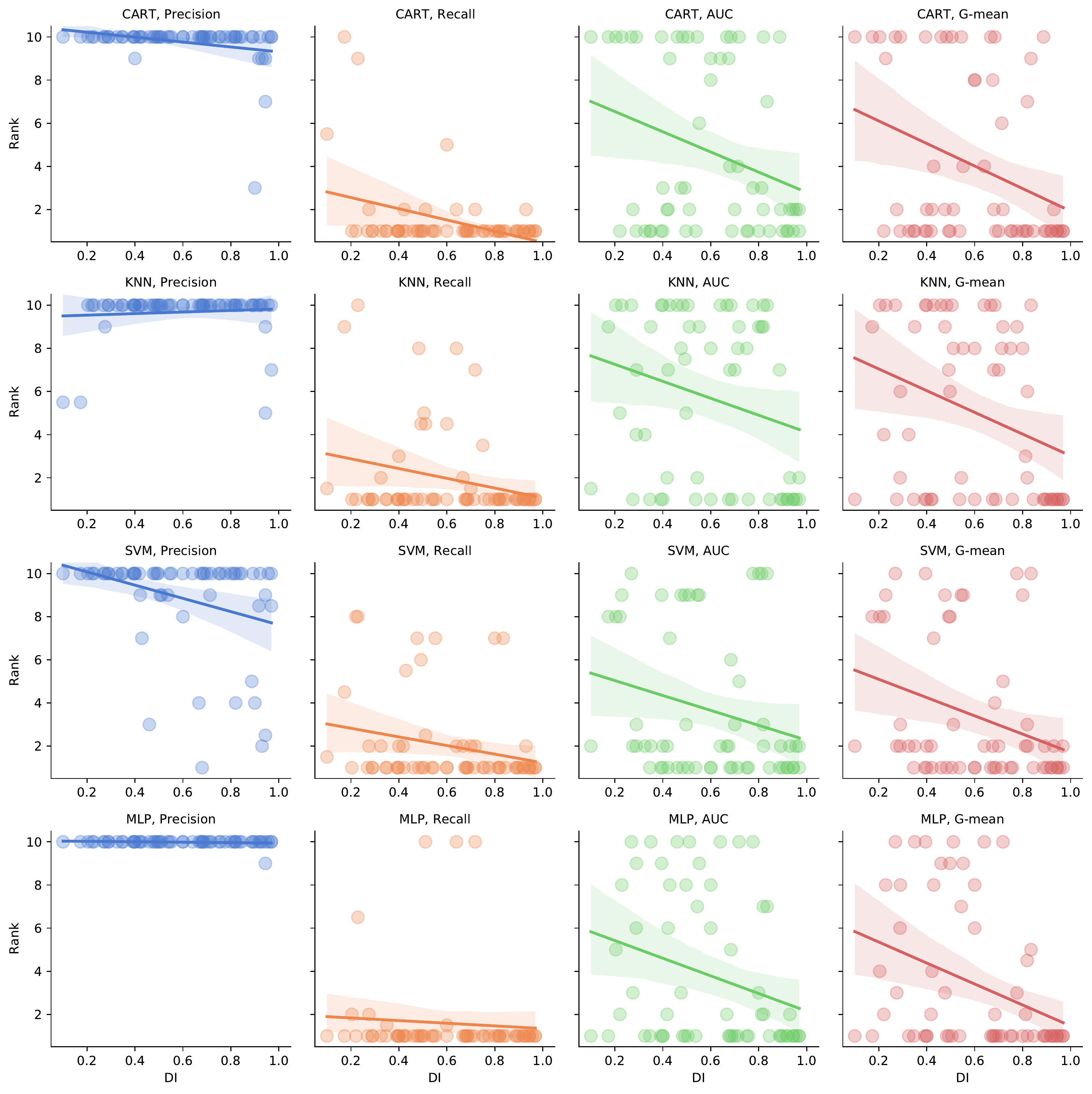}
\caption{Scatterplots representing relation between the data difficulty index of a given dataset and the rank of the PA algorithm achieved on that dataset. 95\% confidence intervals were shown.}
\label{fig:di-impact}
\end{figure*}

\begin{figure*}[!htb]
\centering
\begin{subfigure}[b]{0.4\textwidth}
  \includegraphics[width=\textwidth]{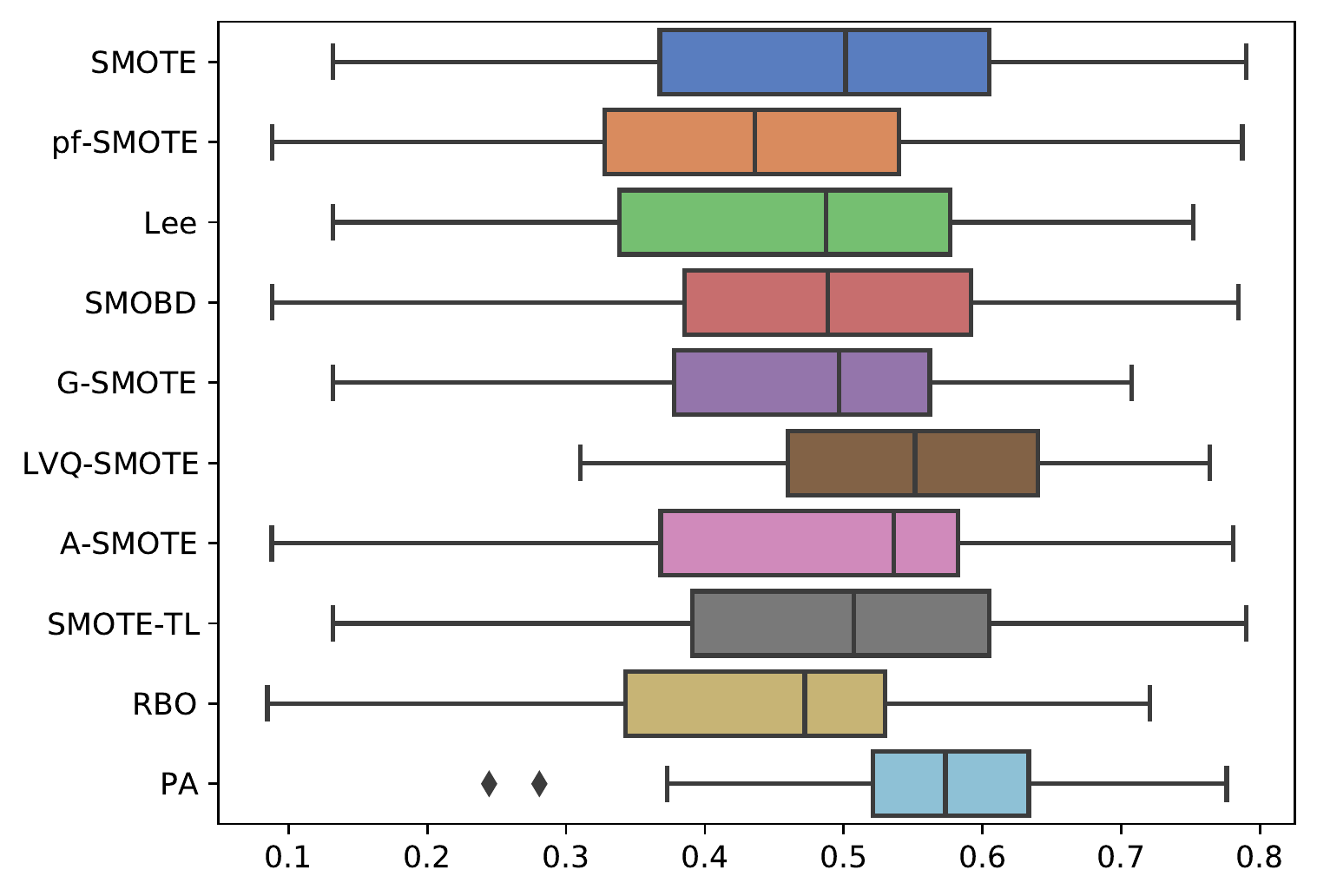}
  \caption{CART}
\end{subfigure}
~
\begin{subfigure}[b]{0.4\textwidth}
  \includegraphics[width=\textwidth]{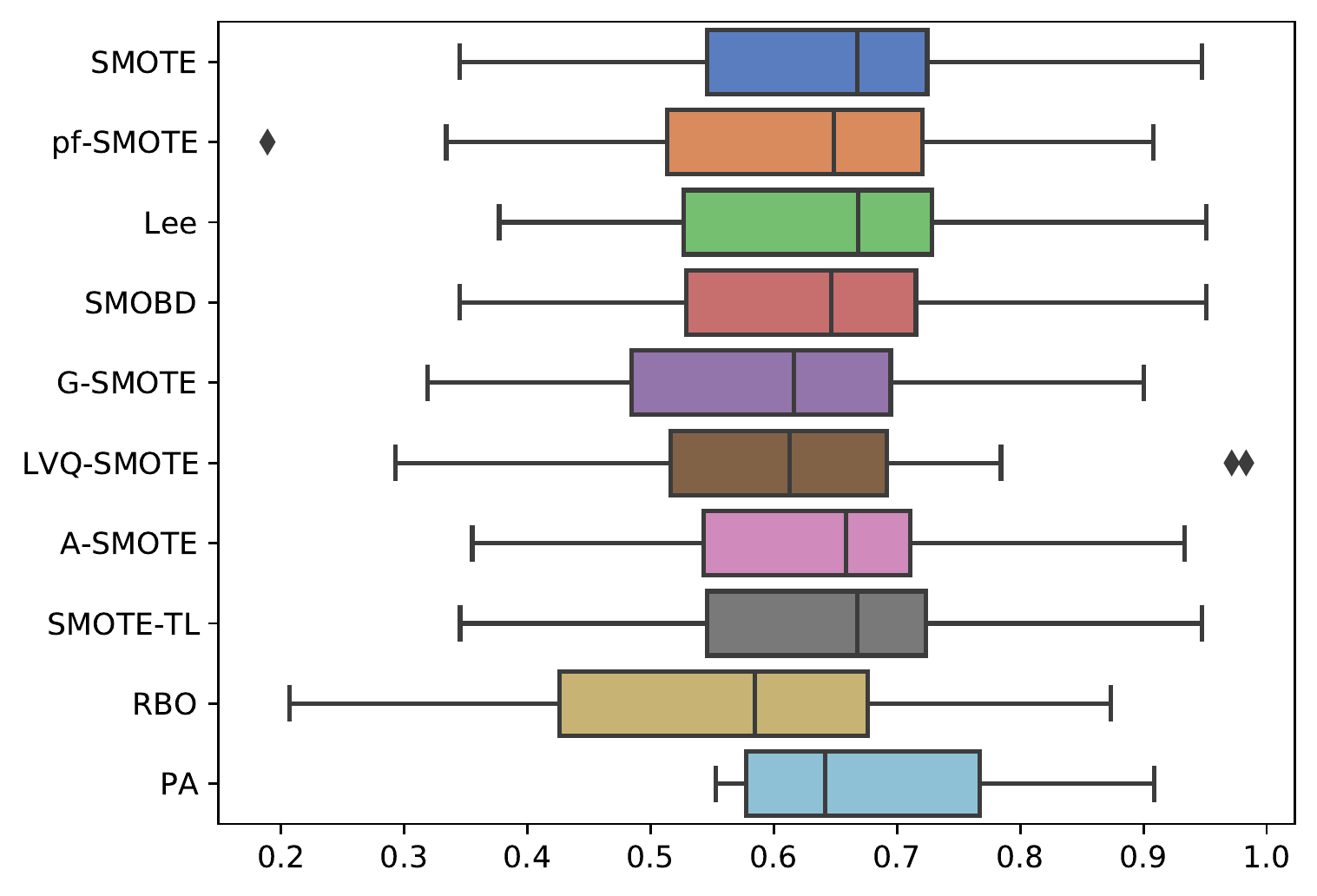}
  \caption{KNN}
\end{subfigure}

\begin{subfigure}[b]{0.4\textwidth}
  \includegraphics[width=\textwidth]{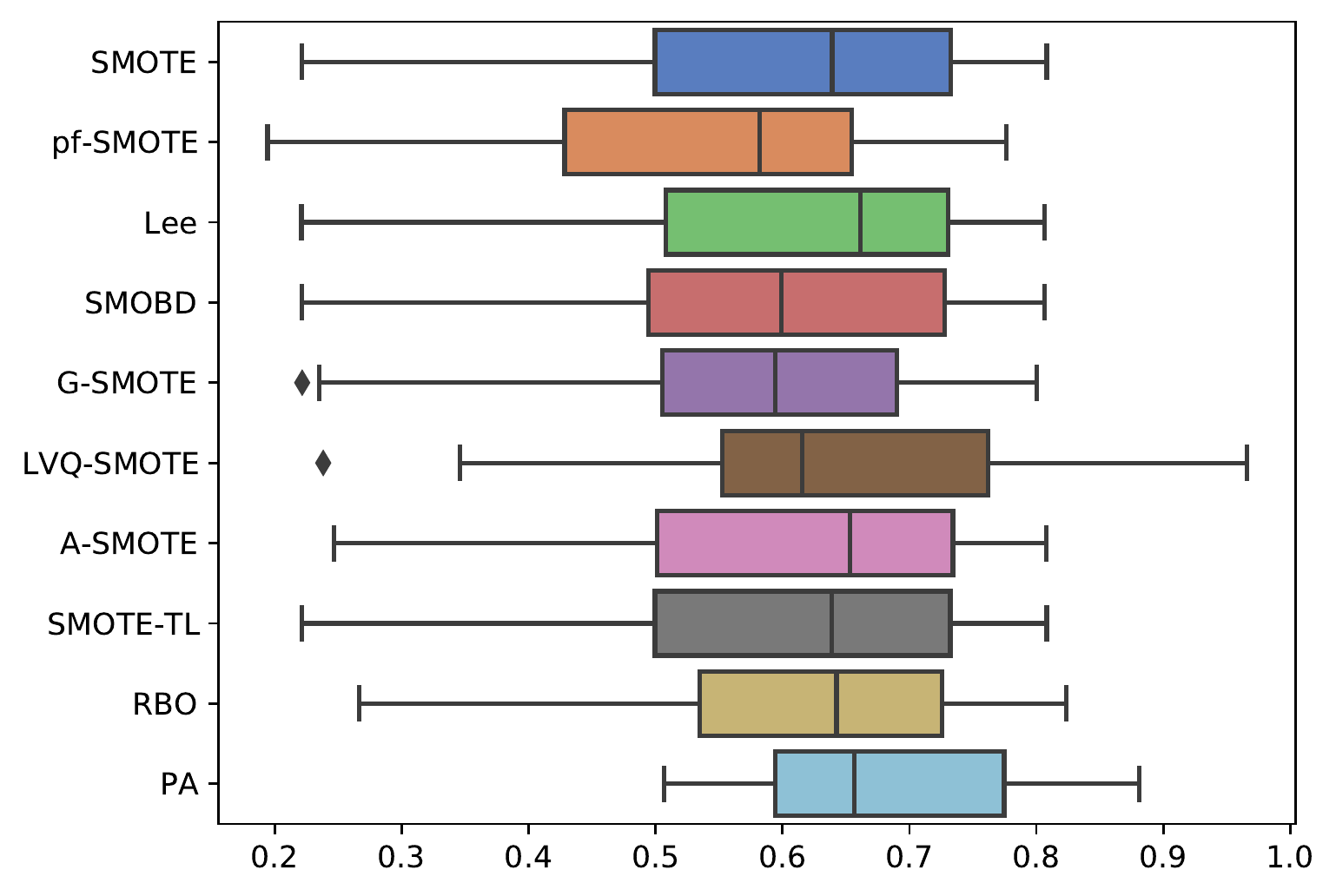}
  \caption{SVM}
\end{subfigure}
~
\begin{subfigure}[b]{0.4\textwidth}
  \includegraphics[width=\textwidth]{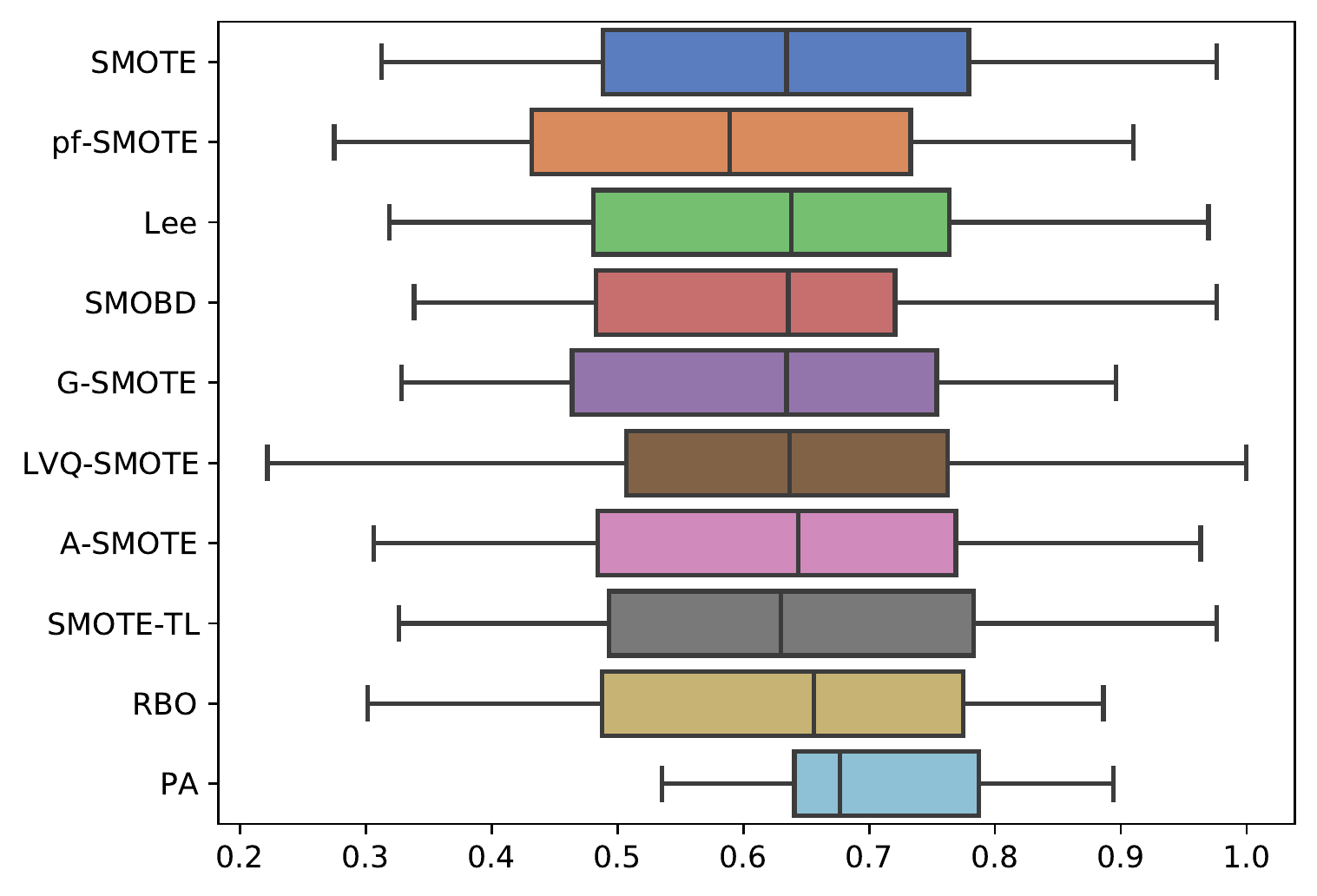}
  \caption{MLP}
\end{subfigure}
\caption{Comparison of average performance of different resampling strategies with respect to G-mean, calculated for the subset of difficult datasets ($DI \geq 0.6$).}
\label{fig:final-boxplots-0.6}
\end{figure*}

Finally, in the last stage of the conducted experiments we tried to examine to what exactly the outperformance of PA can be attributed. To this end we conducted two experiments. In the first of them we examined the relation between the dataset characteristics and relative performance observed for PA. Specifically, we used data difficulty index (DI) to measure the complexity of a given dataset, and correlated it with the rank achieved by PA on that dataset. We present the Pearson correlation coefficients of these two variables in Table~\ref{table:di-corr}, and the scatterplots illustrating this relationship in Figure~\ref{fig:di-impact}. As can be seen, there is a statistically strong correlation between DI and relative performance of PA. This trend is consistent across the classification algorithms and applies to both AUC and G-mean. Interestingly, in contrast to the results presented in Section~\ref{sec:reference}, for two of the considered classifiers a statistically significant improvement in both relative precision and recall was observed for datasets with higher DI. This means that for the more complex datasets not only recall remained higher, but the precision also started to improve (relative to reference methods). Overall, the observed results indicate that PA is particularly well-suited for being applied to complex datasets. This was further illustrated in Figure~\ref{fig:final-boxplots-0.6}, where we once again presented boxplots illustrating the performance of individual methods, this time on a subset of 31 datasets for which DI was equal to or higher than 0.6 (a median DI value observed for all of the considered datasets). As can be seen, the differences between PA and the reference methods are more clearly pronounced compared to those presented previously for all of the datasets in Figure~\ref{fig:final-boxplots-all}.

\begin{table}
\caption{Pearson correlation coefficients between the data difficulty index of a given dataset and the rank of the PA algorithm achieved on that dataset. Statistically significant correlations denoted with bold font. Note that negative correlation indicates increasing performance.}
\label{table:di-corr}
\centering
\begin{tabular}{lllll}
\toprule
 & CART & KNN & SVM & MLP \\
\midrule
Precision & \textbf{-0.2775} & +0.0803 & \textbf{-0.3162} & -0.1876 \\
Recall & \textbf{-0.3712} & \textbf{-0.2494} & \textbf{-0.2376} & -0.0719 \\
AUC & \textbf{-0.2921} & \textbf{-0.2483} & \textbf{-0.2473} & \textbf{-0.2850} \\
G-mean & \textbf{-0.3305} & \textbf{-0.3149} & \textbf{-0.3161} & \textbf{-0.3510} \\
\bottomrule
\end{tabular}
\end{table}

While we were able to show that there is a relation between the complexity of a dataset and relative performance of PA compared to the reference methods, it is not clear to what type of adversity PA is actually resilient. Broadly speaking, we can consider two causes for data complexity. The first one is natural, where complicated decision boundary is required to properly discriminate the data, but the existing observations faithfully describe the underlying class distribution. The second one is artificial, where complexity is caused by a high level of noise, occurring due to factors such as measurement errors, labeling errors, etc., and the observed data does not properly represent real class distribution. It is not clear which of these two factors are the source of data complexity in the considered benchmark datasets.

\begin{figure*}
\centering
\includegraphics[width=0.7\textwidth]{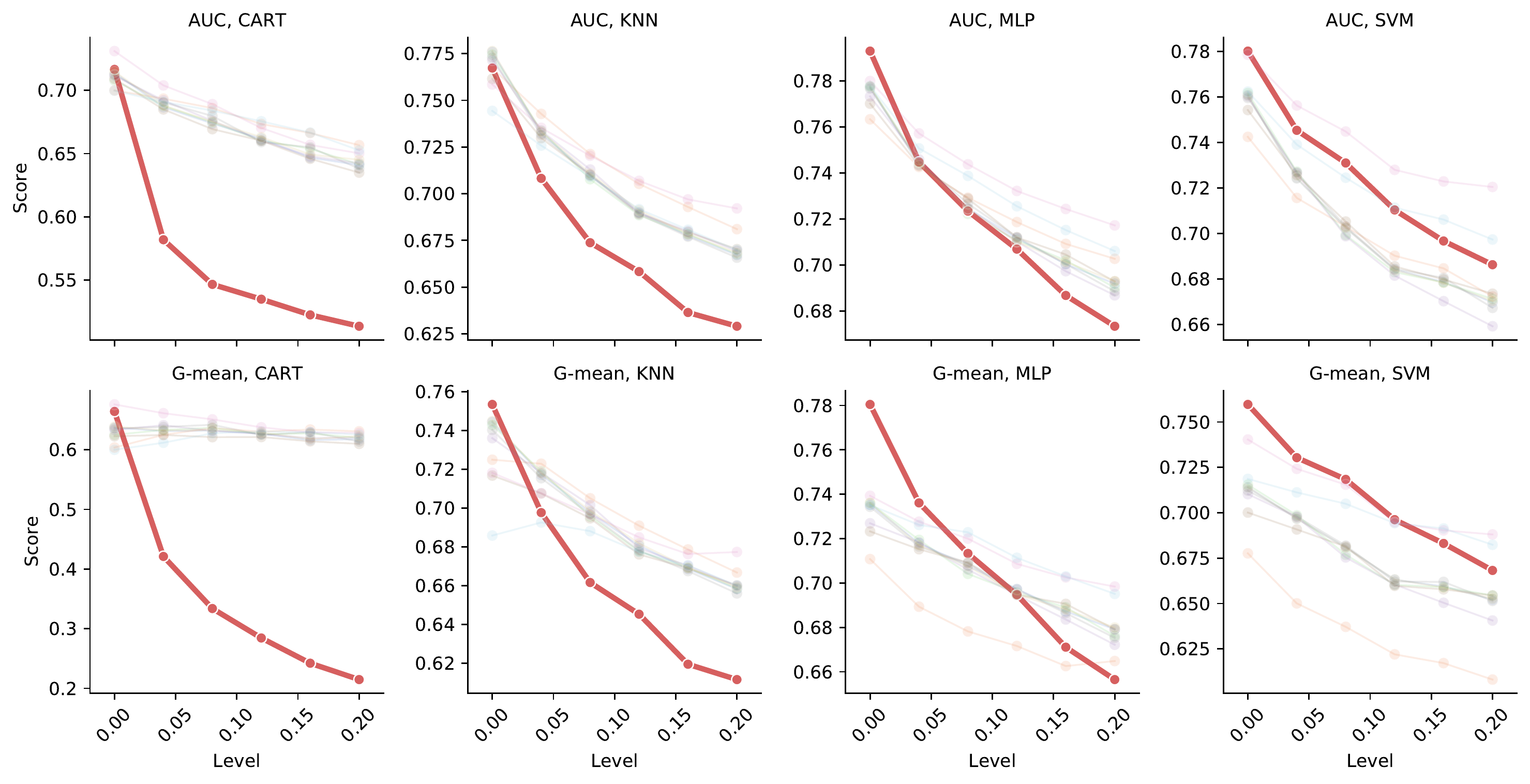}
\caption{Relation between the label noise level and average performance. PA was displayed in the front in red color, and the reference methods were displayed in the background.}
\label{fig:noise-level}
\end{figure*}

To evaluate the impact of the second of the aforementioned sources of complexity we conducted an experiment, during which we artificially introduced noise to the original datasets. Specifically, we considered the case of label noise, in which we randomly switched the label of the original majority class observation to the minority class with a probability equal to the set noise level. During this experiment we modulated the values of noise level $\in \{0.0, 0.04, 0.08, ..., 0.2\}$ and recorded the average performance of different resampling methods. We present the observed results in Figure~\ref{fig:noise-level}. As can be seen, contrary to what could be assumed based on the previously observed relation between data complexity and relative performance, PA does not display resilience to noise. In fact the opposite is true - PA is highly susceptible to the presence of noise, with average performance dropping at a rate significantly higher than the reference methods. This trend was particularly noticeable in the case of CART and KNN classifiers, which can be a possible explanation for a relatively better performance displayed by PA in combination with SVM and MLP, as described in Section~\ref{sec:reference}. Overall, the observed results indicate that while PA seems to outperform other methods on a naturally complex datasets, it is at the same time very prone to the presence of noise. This suggests a rule of thumb for the practical use, according to which it is not recommended to use PA on datasets with known presence of noise (i.e. datasets the labeling of which is highly subjective or prone to error). Furthermore, it suggests that combining PA with other types of data preprocessing, in particular noise removal, might be a feasible direction for further research.

\subsection{Lessons learned}

Based on the presented experimental results we can now attempt to answer the research questions asked at the beginning of this section.

\noindent \textit{RQ1: How do PAO and PAU compare with the previously proposed radial-based resampling strategies?}

\noindent Potential anchoring approach significantly outperformed the previously proposed radial-based strategies, both in the form of over- and undersampling, with stronger differences observed in the case of oversampling. This indicates the usefulness of the proposed approach of preserving the potential shape, and shows that the demonstrated outperformance of PA is not only due to combining the over- and undersampling, but also due to the individual usefulness of both PAO and PAU.

\noindent \textit{RQ2: Is it possible to improve the individual performance of PAO and PAU by combining over- and undersampling?}

\noindent In the conducted experiments we achieved the best performance by combining both over- and undersampling, with the imbalance eliminated in a small proportion by oversampling and a large proportion by undersampling. This trend was consistent across the classification algorithms and the performance metrics, regardless of whether the over- or undersampling achieved a better stand-alone performance in a particular case. It is worth mentioning that this trend was consistent with the results presented in previous studies \cite{koziarski2020csmoute}, indicating that it might be generally applicable to the resampling algorithms.

\noindent \textit{RQ3: How does PA compare with state-of-the-art resampling strategies?}

\noindent PA outperformed the considered resampling algorithms, in particular when combined with either SVM or MLP classifier, for which the proposed method achieved highest average ranks and statistical significance in comparison with a majority of the considered methods. This was achieved by obtaining a significantly better recall of the predictions at the expense of their precision.

\noindent \textit{RQ4: Under what conditions does PA outperform other resampling algorithms?}

\noindent PA achieved the best performance, relative to the reference methods, on difficult datasets, indicating that the proposed approach of preserving the potential shape is particularly well suited for handling complex data. However, our experiments also indicate that PA is as the same time particularly susceptible to the presence of noise, which significantly reduces its performance. Since presence of noise can be difficult to differentiate from natural data complexity without 
a priori knowledge about problem domain, this can pose a challenge for a practical applicability of the method.

\section{Conclusions}
\label{sec:conclusions}

In this paper we proposed a novel approach for handling data imbalance in a manner that preserves the shape of the original class distribution. The proposed approach was utilized in both over- and undersampling in a unified framework. Furthermore, we proposed a measure of imbalanced datasets complexity, which was later utilized to identify the areas of applicability of the proposed approach. The results of our experiments indicate that Potential Anchoring outperforms the considered state-of-the-art resampling strategies, in particular in combination with SVM and MLP classifiers. This outperformance is strongest on a naturally complex dataset. At the same time, however, PA was shown to be susceptible to the presence of noise.

A promising direction for further research is developing mechanisms for reducing the negative impact of noise on the algorithms performance. This can include strategies of pre- and post-processing the data to remove the suspicious observations, both original and generated by PA. Alternatively, a modification to the proposed potential resemblance function can also be utilized. Finally, another research direction worth considering is translating the approach to a big data domain: due to the fact that the approach uses gradient-based algorithms for the optimization, it is feasible to conduct the optimization in a batch mode. Further research on how it would affect the performance is, however, necessary.

\section*{Acknowledgments}

This work was supported by the Polish National Science Center under the grant no. 2017/27/N/ST6/01705 as well as the PLGrid Infrastructure.

\bibliographystyle{elsarticle-num}
\bibliography{main}

\end{document}